\documentclass[nonacm,acmlarge,authorversion]{acmart}

\usepackage[utf8]{inputenc} 
\usepackage[T1]{fontenc}    
\usepackage{url}            
\usepackage{booktabs}       
\usepackage{amsfonts}       
\usepackage{amsmath}
\usepackage{bbm}
\usepackage{nicefrac}       
\usepackage{color}
\usepackage{tikz}
\usepackage{pgfplots}
\usepackage{enumitem}

\usepackage{multirow}
\usepackage{savefnmark}
\usepackage{array}
\usepackage{tabularx}
\usepackage[ruled,vlined,linesnumbered]{algorithm2e}
\usepackage{etoolbox}
\usepackage{graphicx,subcaption}
\usepackage{arydshln}
\usepackage{footnote}

\makeatletter
\patchcmd\algocf@Vline{\vrule}{\vrule \kern-0.4pt}{}{}
\patchcmd\algocf@Vsline{\vrule}{\vrule \kern-0.4pt}{}{}
\makeatother

\makeatletter
\def\adl@drawiv#1#2#3{%
           \hskip.5\tabcolsep
           \xleaders#3{#2.5\@tempdimb #1{1}#2.5\@tempdimb}%
                   #2\z@ plus1fil minus1fil\relax
           \hskip.5\tabcolsep}
\newcommand{\cdashlinelr}[1]{%
     \noalign{\vskip\aboverulesep
              \global\let\@dashdrawstore\adl@draw
              \global\let\adl@draw\adl@drawiv}
     \cdashline{#1}
     \noalign{\global\let\adl@draw\@dashdrawstore
              \vskip\belowrulesep}}
\makeatother
\newcolumntype{L}[1]{>{\raggedright\let\newline\\\arraybackslash\hspace{0pt}}m{#1}}
\newcolumntype{C}[1]{>{\centering\let\newline\\\arraybackslash\hspace{0pt}}m{#1}}
\newcolumntype{R}[1]{>{\raggedleft\let\newline\\\arraybackslash\hspace{0pt}}m{#1}}

\acmBooktitle{}

\definecolor{g-blue}{HTML}{2E86C1}
\definecolor{g-red}{HTML}{B03A2E}
\definecolor{g-purple}{HTML}{AF7AC5}

\usepackage{array}
\newcolumntype{H}{>{\setbox0=\hbox\bgroup}c<{\egroup}@{}}

\newcommand{\tsc}[1]{\textsuperscript{#1}} 

\title{Multi-Stage Conversational Passage Retrieval: An Approach to Fusing Term Importance Estimation and Neural Query Rewriting}

\author{Sheng-Chieh Lin,\tsc{1,2*} Jheng-Hong Yang,\tsc{1,2*} Rodrigo Nogueira,\tsc{1}\\Ming-Feng Tsai,\tsc{3} Chuan-Ju Wang,\tsc{2} and Jimmy Lin\tsc{1}}
\thanks{\tsc{*} Contributed equally.}

\affiliation{\\
\tsc{1} David R. Cheriton School of Computer Science, University of Waterloo\\
\tsc{2} Research Center for Information Technology Innovation, Academia Sinica\\
\tsc{3} Department of Computer Science, National Chengchi University\\[1ex]
}

\renewcommand{\shortauthors}{Lin and Yang, et al.}

\begin{document}

\begin{abstract}
Conversational search plays a vital role in conversational information seeking. 
As queries in information seeking dialogues are ambiguous for traditional ad-hoc information retrieval (IR) systems due to the coreference and omission resolution problems inherent in natural language dialogue, resolving these ambiguities is crucial. 
In this paper, we tackle conversational passage retrieval (ConvPR), an important component of conversational search, by addressing query ambiguities with query reformulation integrated into a multi-stage ad-hoc IR system.
Specifically, we propose two conversational query reformulation (CQR) methods: (1) term importance estimation and (2) neural query rewriting.
For the former, we expand conversational queries using important terms extracted from the conversational context with frequency-based signals.
For the latter, we reformulate conversational queries into natural, standalone, human-understandable queries with a pretrained sequence-to-sequence model.
Detailed analyses of the two CQR methods are provided quantitatively and qualitatively, explaining their advantages, disadvantages, and distinct behaviors.
Moreover, to leverage the strengths of both CQR methods, we propose combining their output with reciprocal rank fusion, yielding state-of-the-art retrieval effectiveness, 30\% improvement in terms of NDCG@3 compared to the best submission of TREC CAsT 2019.
\end{abstract}

\maketitle

\section{Introduction}

The fundamental purpose of conversational search systems is to provide relevant results that satisfy users' information needs through natural language communication~\cite{conv_search,SWIRL2018}.
Whereas both researchers and practitioners have made great progress in dialogue modeling~\cite{cqu,CANARD}, there is still a large gap between automatic systems and oracle queries formulated by human annotators in terms of retrieval effectiveness~\cite{cast,quretec,vakulenko2020question,yu2020fewshot}.
In this paper we seek to bridge this gap by combining term importance estimation and neural query rewriting that emulates the behavior of human annotators.
In particular, we seek to demonstrate the three following claims:

\begin{enumerate}
    \item Integrating conversational query reformulation modules in multi-stage ad-hoc retrieval systems is practical.
    \item Query variations from our proposed term importance estimation and neural query reformulation techniques improve retrieval metrics in markedly different ways.
    \item Fusing query variations that distinctly reflect users' information needs improves retrieval effectiveness.
\end{enumerate}
Conversational search plays an essential role in conversational information seeking~\cite{conv_search}, which involves two states defined in~\citet{penha2019mantis}:\ (1) information-need elucidation and (2) information presentation.
For the former, the goal is to help users understand, clarify, refine, express, and elicit their information needs, whereas the latter is to retrieve and present relevant information in a conversational manner.
In this paper, we focus on tackling conversational passage retrieval (ConvPR), an important component of conversational search, by addressing a multifaceted and challenging problem:
How to better elucidate the information needs of users with query reformulation.

Information seeking dialogues often contain colloquial and contextually dependent utterances, the queries\footnote{For convenience, we refer to natural language questions as queries.} of which are often ambiguous for traditional ad-hoc information retrieval (IR) systems.
Although concise expressions of queries are normal during human conversations, such succinct expressions are usually not effective for ad-hoc IR systems due to the coreference and omission resolution problems inherent in natural language dialogue.
One typical solution to address these issues is utilizing a multi-stage system composed of a natural language understanding module for information-need elucidation and a relevant document-collecting module for information presentation~\cite{quretec,vakulenko2020question,yu2020fewshot}.

Most recent studies address the issue of colloquial expressions through neural conversational query rewriting~\cite{cqu, CANARD, vakulenko2020question, lin2020conversational, yu2020fewshot}.
For example, some propose reformulating queries with a neural language model (LM) trained with a sequence-to-sequence (seq2seq) neural architecture using rewritten queries by humans~\cite{CANARD, cqu}.
Such a conversational LM thus learns to generate a nearly coreference-and-omission-free query at each turn, which is conditioned on both the original ambiguous query and queries from previous turns in the conversation history.
Because previous work~\cite{cqu, CANARD, lin2020conversational} on conversational query rewriting (or reformulation) evaluates model effectiveness using intrinsic metrics of text similarity, the impact on downstream retrieval systems is not directly assessed.
In contrast, this work, like that of~\citet{vakulenko2020question} and~\citet{yu2020fewshot}, directly evaluates retrieval effectiveness using extrinsic metrics.

\begin{figure}[!h]
\begin{subfigure}{.49\columnwidth}
    \includegraphics[width=\columnwidth]{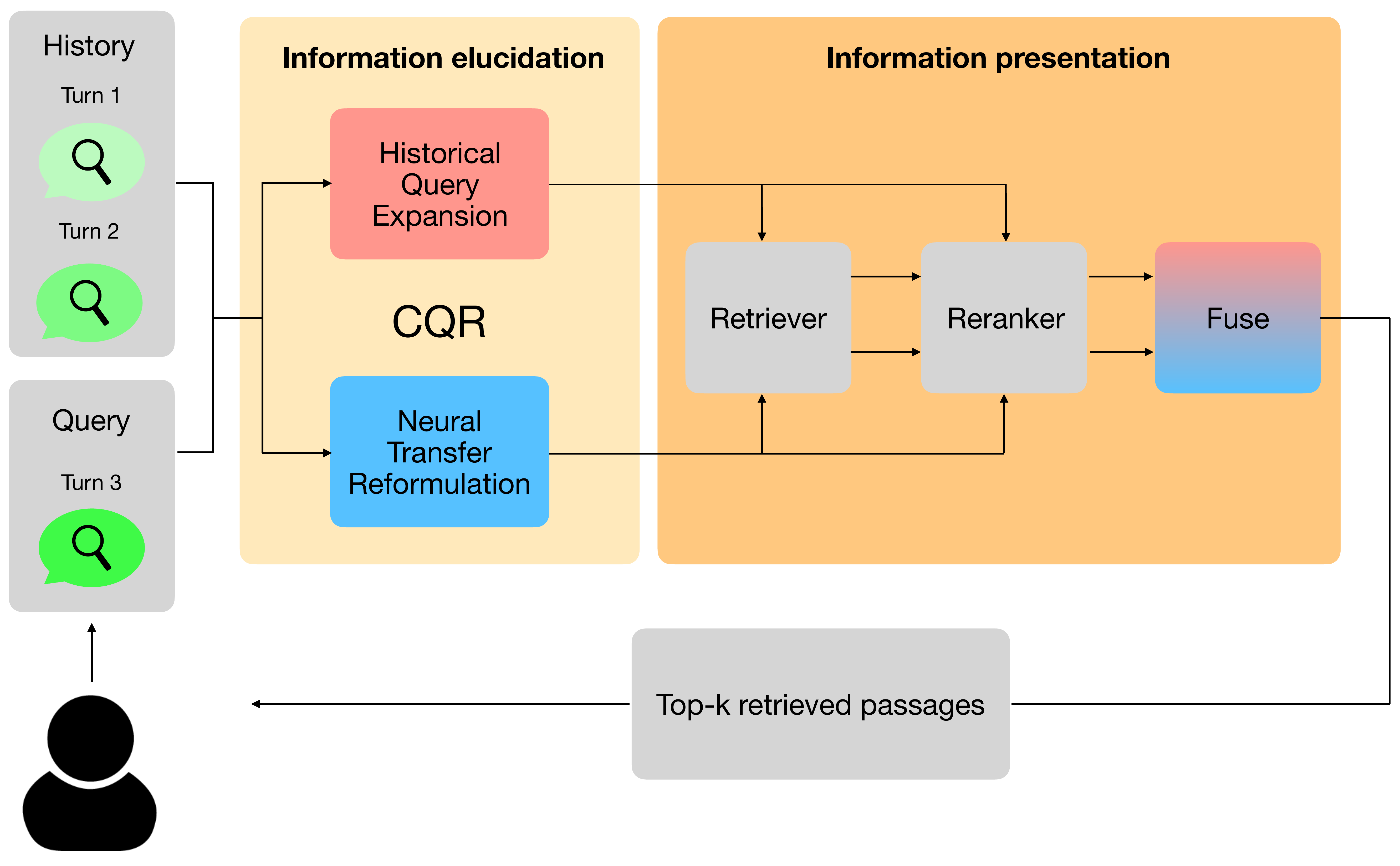}
    \caption{Late fusion of query variations}
\end{subfigure}
\begin{subfigure}{.49\columnwidth}
    \includegraphics[width=\columnwidth]{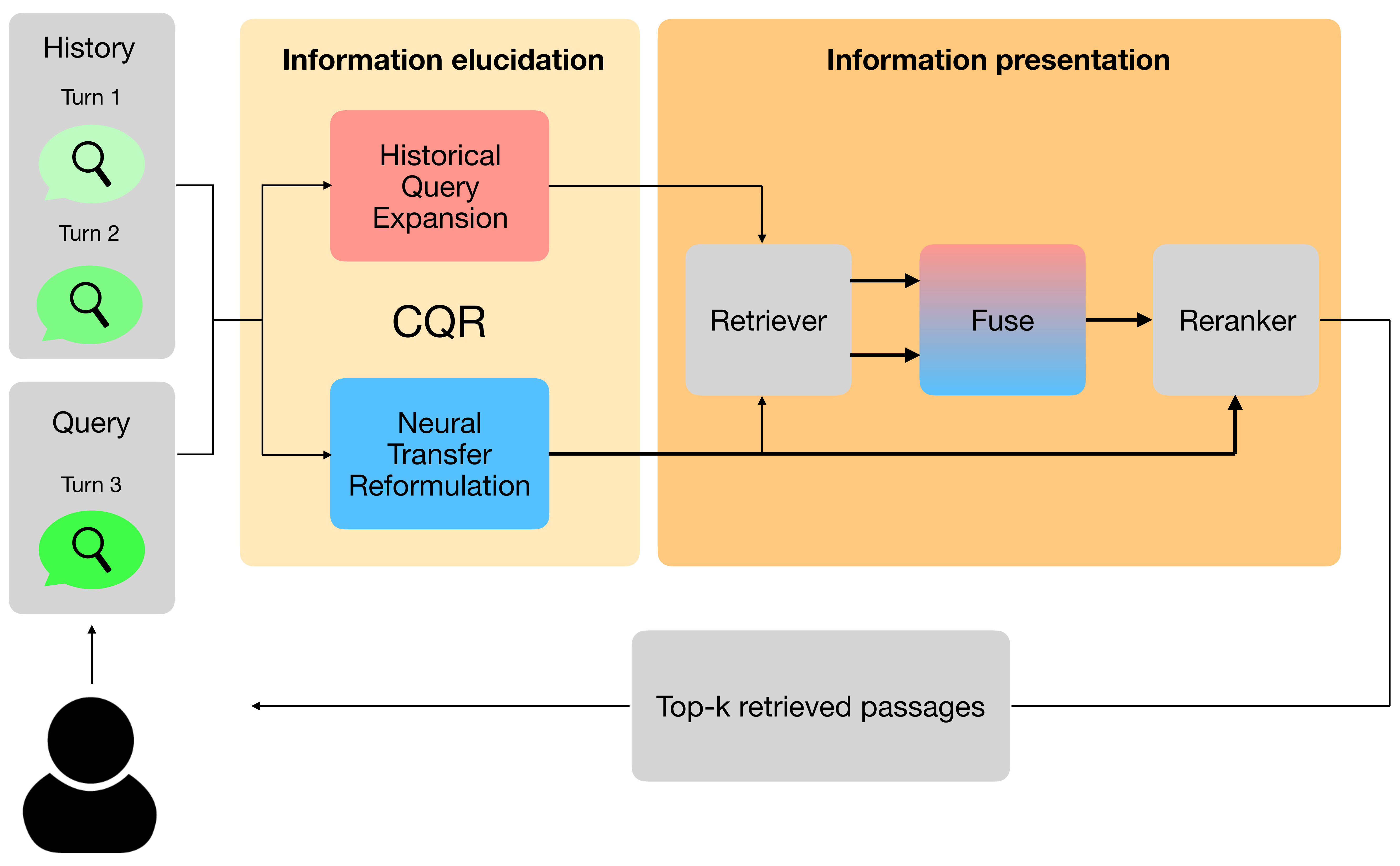}
    \caption{Early fusion of query variations}
\end{subfigure}
\caption{Two possible architectures for multi-stage conversational passage retrieval. Panel (a): two types of queries run in parallel through the pipeline, and their re-ranked lists are fused as the final output.
Panel (b): ranked lists are fused at the first-stage BM25 retrieval, and then both the fused list and the neural rewritten query serve as inputs for BERT re-ranking.}
\label{fig:configuration}
\end{figure}

Quantifying the term importance of each query or document, in turn, is a key ingredient of state-of-the-art ad-hoc IR systems~\cite{deepct}.
Traditional bag-of-words (BOW) based systems, such as TF-IDF and BM25~\cite{bm25}, use frequency-based signals to estimate relevance between queries and documents.
Conventional ConvPR pipelines (see Figure~\ref{fig:configuration}) often involve an ad-hoc IR system; in this scenario, the estimation of term importance may be important for information-need elucidation.
However, this insight is under-explored in ConvPR in the literature and under-utilized as a part of a conversational query reformulation module.
Coreferences (or omissions) in multi-turn dialogues exacerbate query ambiguity by replacing (or deleting, respectively) important terms that have strong frequency-based signals, which are often crucial for clarifying information needs.
Although the aforementioned conversational LMs do manage to account for ambiguous queries by restoring some missing informative terms in the queries, the issue of term importance has yet to be satisfactorily addressed in our opinion.

Combining query variations that represent the same information needs is an effective technique for ad-hoc information retrieval~\cite{belkin1993,Benham2019boost}.
However, little has been done to study query variations from conversational query reformulation systems.
Previous studies indicate that different users pose different queries to represent the same information need~\cite{noreault1980,belkin1993}.
Similarly, in the context of conversation understanding, different conversational query reformulation models could interpret users' information needs from different perspectives.
In this paper, we conduct an in-depth analysis of how different query reformulation models represent users' information needs and the impact they have on retrieval effectiveness.
While most previous studies attempt to deliver a single reformulated query per information need in a conversational search system~\cite{cqu, CANARD, lin2020conversational,cast,quretec,vakulenko2020question,yu2020fewshot}, here we consider term importance estimation and neural query rewriting to represent a single information need for two purposes.
First, term importance signals are essential for traditional BOW-based models~\cite{deepct}.
Second, linguistic structures in natural language queries generated by neural query rewriting play a vital role for the contextual neural language model BERT~\cite{BERT} in text ranking~\cite{dai2019deeper}.
Thus, given these two considerations, we use queries from multiple reformulation modules and study their retrieval effectiveness with both a BOW-based model and a BERT-based model.

To address the multifaceted challenges of ConvPR, we propose adapting and aggregating existing NLP/IR techniques for information-need elucidation and information presentation~\cite{penha2019mantis}.
First, we address information-need elucidation with query reformulation using both term importance estimation and neural conversational query rewriting, for which we develop two conversational query reformulation (CQR) modules.
For term importance estimation, inspired by the nature of topic-orientated information seeking dialogues and term importance estimation through frequency-based signals, we propose a technique called Historical Query Expansion (HQE) to clarify query ambiguities and to quantify the importance of informative terms in a query through frequency-based signals (e.g., BM25 scores) from conversation history.
For query rewriting, we present Neural Transfer Reformulation (NTR), an approach that builds on the previous work of~\citet{lin2020conversational, vakulenko2020question} to fine-tune the pretrained Text-to-Text Transfer Transformer (T5)~\cite{t5}.
Next, we build a conversational multi-stage retrieval system using the proposed CQR modules, a multi-stage ad-hoc IR system~\cite{marco_BERT}, and a fusion module based on reciprocal rank fusion~\cite{rrf} to aggregate the query variations from the CQR modules.
To further improve ConvPR effectiveness, we study the characteristics of the query variations from the different QR modules and propose multi-stage configurations that integrate their results via reciprocal rank fusion.
Possible architectures for this integration are illustrated in Figure~\ref{fig:configuration}.
Specifically, we analyze both late and early fusion of query variations:
In the former, we apply queries reformulated from CQR modules to both a BOW retriever and a BERT-based re-ranker separately, after which we fuse their final top-$k$ retrieved passages, as shown in Figure~\ref{fig:configuration}(a).
In the latter, we perform early fusion at the retrieval stage followed by a re-ranking stage that takes only the queries from a neural query rewriting module, as shown in Figure~\ref{fig:configuration}(b).

Evaluating complex conversational search systems is a challenging research problem.
Fortunately, we are able to provide a practical evaluation of ConvPR under the framework built by the Conversational Assistant Track (CAsT) hosted by the Text REtrieval Conference (TREC) 2019~\cite{cast}.
We evaluate, in an end-to-end manner, the two types of reformulated queries from the proposed CQR models and their overall effectiveness through commonly adopted metrics such as mean average precision (MAP) and normalized discounted cumulative gain (NDCG@$k$).
Furthermore, we conduct analyses to compare the results of different types of queries, including raw queries, reformulated queries from the two modules, and queries annotated by humans, for both BOW-based retrieval and BERT-based re-ranking.
Furthermore, we leverage BERT-based embeddings, the Jaccard similarities of retrieved sets, and BLEU scores~\cite{BLEU} to study information-need elucidation in ConvPR.
Finally, we demonstrate the effectiveness of our proposed fusion techniques and conclude by attributing the success of the fusion of query variations to (1) the ability of both CQR methods to track the topics of conversation across turns, (2) query term importance re-weighting from HQE, and (3) the preservation of linguistic structures in queries from NTR.

The contributions of this work are summarized as follows:
\begin{itemize}
	\item We demonstrate the effectiveness of two conversational query reformulation approaches (HQE and NTR) stacked on top of a widely-used multi-stage search architecture.
	
	\item We conduct detailed analyses of the HQE and NTR approaches quantitatively and qualitatively, explaining their advantages and disadvantages. 
	One variant of HQE was the best automatic submission to TREC CAsT 2019; NTR further improves on this by 16\% in NDCG@3.
	
	\item We demonstrate that fusing results from the two CQR modules yields the highest effectiveness for conversational passage retrieval.
\end{itemize}
In sum, this work demonstrates how to tackle ConvPR using term importance estimation with BM25 and conversational query rewriting using a pretrained Transformer-based seq2seq model, which we combine to build a simple yet effective rank fusion pipeline for ConvPR in conversational search.

\section{Related Work}

\textbf{Conversational search.}
Conversational search~\cite{penha2019mantis} seeks to facilitate IR in a conversational context, covering two main perspectives:\ information-need elucidation and information presentation.
Information-need elucidation focuses on understanding users' information needs in dialogue contexts, such as query clarification~\cite{Aliannejadi2019AskingCQ}, suggestion~\cite{q_suggest}, and disambiguation~\cite{cqu, CANARD}.
On the other hand, information presentation covers a broad range, such as recommendation, information retrieval, and question answering~\cite{sun-etal-2018-open, wang2018evidence}. 
Many recent studies concern the combination of the two research topics.
For example, multi-task training strategies have been proposed that combine query suggestions and document retrieval~\cite{conv_search, uddin2018multitask, uddin2019context}.
However, these works neither tackle query disambiguation explicitly nor consider information needs in a ConvPR task---that is, a task that requires a system to search a corpus of passages using natural language queries from a conversation.
For this specific task, \citet{cast} provide a conversational search dataset for the Conversational Assistance Track (CAsT) at TREC~2019, seeking to measure system effectiveness in open-domain information retrieval while serving users' information needs expressed in colloquial dialogues.
Thus, utterances include common natural language features beyond keyword queries, e.g., coreference, omission, paraphrasing, ambiguous intents, etc.
The task challenges participants to tackle query disambiguation and open-domain information retrieval in a combined manner.

Note that in the literature, three papers~\cite{ilips,harness,quretec} are closely related to our study, in the sense that they also build query reformulation modules on top of multi-stage IR systems.
In the first work, \citet{ilips} propose unsupervised query expansion using keywords from conversational contexts, in which keywords are extracted according to a word graph built through external data (i.e., word embeddings). 
The second work by~\citet{harness}, in contrast, expands queries by teaching models to optimally concatenate previous utterances in conversational contexts.
Finally, in the third paper, \citet{quretec} formulate query expansion as a term classification task and further improve model effectiveness with distantly supervised learning.
Note that both \citet{quretec, harness} use encoder-only models, which directly concatenate terms or previous utterances into current queries without considering linguistic structures in the natural language queries.
In contrast to these papers, we propose an unsupervised query expansion model with a keyword extractor based on term importance estimation inspired by relevance models~\cite{rlm2001}, together with a supervised query reformulation method using sequence-to-sequence models to retain linguistic structures in natural language.
In addition, we study the characteristics of both proposed methods and find that they complement each other in terms of retrieval effectiveness.
We then propose a fusion pipeline and empirically validate its effectiveness.

\smallskip \noindent
\textbf{Multi-stage retrieval systems.}
Multi-stage retrieval systems are comprised of a candidate generation process followed by one or more re-ranking stages to strike a balance between effectiveness and efficiency~\cite{Jimmy2013, Nicola2013, Clarke2016}.
Relevant research includes work on feature extraction efficiency~\cite{docvec_multistage}, dynamic cutoff depth~\cite{cutoff_pred}, shard prediction~\cite{shard_pred}, and joint cascade ranking optimization~\cite{cascade, multi_neural, marco_BERT, birch, multi_BERT}.
The foundation of our work is built on a competitive cascade pipeline proposed by~\citet{marco_BERT} and~\citet{birch}:\ BM25 candidate generation followed by BERT re-ranking, the effectiveness of which has been demonstrated in representative IR datasets such as Robust04, TREC CAR, and MS MARCO~\cite{voorhees2004overview,car,marco}.

\smallskip \noindent
\textbf{Query Reformulation (QR).}
QR has proven effective in IR.
For example, \citet{qe1} and \citet{rm3} expand a query with terms from retrieved documents; \citet{rl-query_reform} exploit reinforcement learning for query reformulation.
These techniques are also widely used in session search~\cite{rlm_session, sess_search0}.
Note that although many studies focus on improving the effectiveness of ad-hoc queries, we focus on QR in a conversational context.
Among QR studies, the papers most relevant to ours are~\citet{cqu, CANARD, ir_qr}, all of which demonstrate the feasibility of deep learning for reformulating conversational queries. 
However, they examine only one facet of effectiveness in terms of question-in-context rewriting.
In this work, we apply and analyze a method based on query expansion as well as a method based on transfer learning~\cite{transfer, pre-train} in a full conversational IR pipeline.

\smallskip \noindent
\textbf{Relevance Model (RM).}
RM~\cite{rlm2001} is a method to estimate query term importance by measuring co-occurrences between query terms and terms in the top relevant documents.
RM has a variety of applications, including ad-hoc information retrieval~\cite{rm3}, recommendation~\cite{rlm_rec}, and session search~\cite{rlm_session}.
However, to our best knowledge, we are the first to apply these ideas to CQR and demonstrate their effectiveness in a conversational IR pipeline.

\smallskip \noindent
\textbf{Query variations.}
The study of query variations seeks to represent the same information need from different perspectives posed by different searchers.
\citet{noreault1980} show that when different users create queries to address the same information need, the intersection of retrieved documents given their queries is surprisingly low.
While this variation in information needs poses a problem when selecting ``optimal'' queries, combining variants serves as an effective way to improve retrieval quality~\cite{belkin1993}.
In the context of ConvPR, information elucidation modules could also interpret users' information needs in different ways by adopting different modeling perspectives.
Thus, combining queries from different elucidation modules should be considered when designing a multi-stage ConvPR system because various forms of a query may yield different advantages and capitalize on different ranking signals.
For example, linguistic structures in natural language queries are beneficial for BERT-based re-ranking models~\cite{dai2019deeper};
hence, they prefer elucidation models that produce fluent natural language queries.
On the other hand, queries that explicitly capture term importance could benefit traditional BOW-based retrieval models~\cite{deepct}.
More importantly, this paper sheds light on applying query variants to ConvPR.
That is, for researchers and practitioners, how to combine conversational query variants is a crucial design focus when meeting the requirements of a retrieval system.

\section{Problem Setup}

\smallskip \noindent
\textbf{Conversational Passage Retrieval (ConvPR).}
ConvPR is defined as an IR task in a conversational context.
Given a sequence of conversational utterances $u^{s} = (u_{1}, \cdots, u_{i},
u_{i+1}, \cdots )$ for a topic-oriented session $s \in S$, where $S$ is the set
of all dialogue sessions and $u_{i}$ stands for the $i$-th utterance ($i
\in \mathbb{N}^{+}$) in the session (referred to as turn $i$), the goal of this task is to find a set of
relevant passages $\mathcal{P}_{i}$, for each turn's user utterance $u_i$ that satisfies the information needs in turn $i$ given the context in previous turns $u_{<i} = (u_{1}, \cdots, u_{i-1})$.

\smallskip \noindent
\textbf{Task scope.}
To facilitate the ConvPR task and to provide a reusable test collection, the organizers of CAsT in TREC 2019 began with a selection of open-domain exploratory information needs $I$ and provided a predefined set of topic-oriented sessions $S^{I}$.\footnote{\url{http://www.treccast.ai/}}
In addition, a passage collection $\mathcal{C} \supseteq \mathcal{P}_i$ was provided to retrieve candidate responses for each turn in these sessions.

Under the CAsT setting, the utterances in the provided topic-oriented
sessions $S^I$ not only control the complexity of the task but also mimic 
features of ``real'' dialogues via the following properties:

\begin{itemize}
	\item Utterance transitions are coherent between turns in a given
	 topic-oriented session.
	\item Utterances are fluent natural language questions, similar to questions in the Google Natural Questions dataset~\cite{natural_q}.
    \item The dialogues include such natural language features as coreferences and omissions.
    \item Turns depend only on previous utterances and not system responses.
    \item Comparisons between subtopics are introduced.
\end{itemize}

\smallskip \noindent
\textbf{Conversational multi-stage retrieval systems.}
\label{sec:multi-stage}
To reuse existing IR pipelines and benefit from the fine-tuned effectiveness of relevance prediction models, a typical approach for ConvPR is to reformulate user utterances with their context into suitable standalone queries and feed the reformulated queries into the pipelines.

For an IR system, let $P\,(R=1\,|\,q, p)$ denote the probability of relevance conditioned on a query-passage pair $(q,p)$, where $R=1$ denotes that passage $p\in\mathcal{C}$ is relevant to query~$q$ (otherwise, $R=0$).
Currently, one popular approach is to further factorize $P\,(R=1\,|\,q, p)$ into a multi-stage pipeline $f_{\theta} \circ f_{\phi} \propto P(R=1\,|\,q, p)$ to balance effectiveness and efficiency:\
$f_{\phi}$ is a BOW retrieval model such as BM25 and $f_{\theta}$ represents a high-quality re-ranker using neural networks or other machine learning methods.

Likewise, for ConvPR, we factorize the probability of retrieving a relevant passage $p \in \mathcal{P}_i$ for each turn $i$ with an information set $\{u_i$, $u_{<i}\}$ that comprises the utterances by turn $i$ as 
\begin{equation}
\label{eq:multi_stage}
	P\,(R=1\,|\, \{u_i, u_{<i}\}, p) = P\,(R=1\,|\, q_i, p)\,P\,\left(q_i\,|\,\{u_i, u_{<i}\}\right) \text{.}
\end{equation}
With this formulation, ConvPR can be approximated by separately maximizing the probabilities of (a) a relevance prediction model $P\,(R=1\,|\,q_i, p)$ and (b) a query reformulation model $P\,(q_i\,|\{u_i, u_{<i}\})$.
Thus, the goal of a query reformulation model is to reformulate a raw conversational user utterance $u_i$ for each turn $i$ into a clear and informative standalone query $q_i$ for the relevance prediction model~\cite{ir_game}.

As the relevance judgments in the training set from CAsT are sparse and very limited (see Table~\ref{tb:statistics}), we here focus on query reformulation methods and leave the burden of tuning a relevance prediction
model to a known competitive pipeline---BM25 with BERT---in large-scale
passage ranking tasks~\cite{marco, car, marco_BERT}.

\smallskip \noindent
\textbf{Conversational Query Reformulation (CQR).}
The goal of CQR is to produce an informative standalone query~$q_i$ for each turn~$i$ for downstream relevance prediction models.
Specifically, given an information set $\{u_i, u_{<i}\}$ that includes the
utterances by turn $i$, the task of CQR consists of the following two
components:\ (a) filtering out unnecessary information in $\{u_i, u_{<i}\}$ and (b) generating the query $q_i$ from the filtered information.
Thus, with CQR we seek a function $q_i=g\left(\{u_i, u_{<i}\}\right)$, the output of which (i.e., $q_i$) maximizes the probability in Eq.~(\ref{eq:multi_stage}).

However, given the limited number of relevance labels available in existing datasets, using supervised learning to construct a parametric function to maximize Eq.~(\ref{eq:multi_stage}) is difficult.
In this work, we propose two label-free approximations for CQR.
The first we call Historical Query Expansion (HQE):\ a heuristics-based model  $g_{{\phi}}(\cdot)$ (see Section~\ref{sec:HQE}).
The second we call Neural Transfer Reformulation (NTR):\ an off-the-shelf
data-driven neural model $g_{{\theta}}(\cdot)$ pretrained on an out-of-domain dataset under a transfer learning paradigm (Section~\ref{sec:T5}). 
Note that both approaches only approximate $g(\cdot)$ with $g_{{\phi}}(\cdot)$ and $g_{{\theta}}(\cdot)$, respectively, due to the fact that the objective of $g(\cdot)$ in fact involves optimizing the end-to-end effectiveness of a ConvPR system.

\begin{table}[t!]
	\caption{TREC CAsT 2019 Training Topic 1. A conversation session consists of several questions. Each question generated by a user continues previous utterances. The task is to find relevant passages for each question based on its previous utterances.}
	\label{tb:CAsT_example}
	\centering
	\small
    \begin{tabular}{cl}
	\toprule
    \multicolumn{2}{p{8cm}}{\textbf{Title}: Career choice for Nursing and Physician's Assistant}   \\
	\midrule
    Turn ($i$) & Conversation utterances ($u_i$) \\
    \midrule
    1&	What is a physician's assistant? \\
    2&	What are the educational requirements required to become one? \\
    3&	What does it cost? \\
    4&	What's the average starting salary in the UK? \\
    5&	What about in the US? \\
    6&	What school subjects are needed to become a registered nurse? \\
    7&	What is the PA average salary vs an RN? \\
    8&	What the difference between a PA and a nurse practitioner? \\
    9&	Do NPs or PAs make more? \\
    10 &Is a PA above a NP? \\
    11 & What is the fastest way to become a NP? \\
    12&	How much longer does it take to become a doctor after being an NP? \\
	\bottomrule
	\end{tabular}
\end{table}

\section{Methodology}

To develop models for CQR, we start by observing two characteristics of conversational user utterances from the CAsT training data in Table~\ref{tb:CAsT_example}:

\smallskip \noindent
\textbf{Observation \#1: Main topic and subtopic.}
A session is centered around a main topic, and the turns in the session dive
deeper into several subtopics, each of which only lasts a few turns. 
For instance, in Table~\ref{tb:CAsT_example}, the main topic of the session is
``physician's assistant'':\ turns~2 and 3 discuss the
subtopic of ``educational requirements'' while turns~4 and 5 are related to
the subtopic of ``average starting salary''.

\smallskip \noindent
\textbf{Observation \#2: Degree of ambiguity.}
The degree of ambiguity divides utterances into three categories:
The first category includes utterances with clear intents, which can  be directly treated as ad-hoc queries, such as turns~1 and 6 in
Table~\ref{tb:CAsT_example}.
The second category contains those starting a subtopic (e.g., turns~2 and 4), and the last category is composed of ambiguous utterances that
continue a subtopic (e.g., turns~3 and 5). 

\medskip
\noindent Based on the above observations, we propose two CQR methods: (1) Historical Query Expansion (HQE), a heuristic query expansion strategy; (2) Neural Transfer Reformulation (NTR), a data-driven approach transferring human knowledge to neural models from human annotated queries.

\subsection{Historical Query Expansion}\label{sec:HQE}

We first introduce HQE to heuristically capture the above observations.
Specifically, there are three main steps in HQE.
For each utterance in a session, we
(1) extract the main topic and subtopic keywords from the utterance; 
(2) measure the ambiguity of the utterance; 
(3) expand queries for the ambiguous utterances with the main
topic and subtopic keywords extracted from previous turns.
We propose keyword extractor and query performance predictor modules
to carry out these three steps for constructing the function $g_\phi$.

\subsubsection{Keyword Extractor ($KE$)}
Given an utterance $u_i$ consisting of $n(u_i)$ tokens, the utterance is
represented as a tuple $\left(t_i^1, \dots, t_i^{n(u_i)}\right)$,
where $t_i^k$ denotes the $k$-th token in $u_i$.
The aim of the KE is to compute the score of each token in the utterance such
that the score captures the importance of the token in the utterance.
Drawing from relevance models~\cite{rlm2001}, for each token, we propose leveraging the retrieval score of its highest-scoring
document to characterize its importance in the utterance as
\begin{equation*}
\mathcal{R}_i^k=\text{KE}\left(t_i^k, \mathcal{C}\right)=\max\limits_{p \in \mathcal{C}} \left\{ \mathcal{F}_{\text{KE}}\left({t}_i^k,p\right) \right\},
\end{equation*}
where $\mathcal{R}_i^k$ denotes the importance score of token~$t_i^k$, and
$\mathcal{F}_{\text{KE}}(\cdot)$ is the function to compute the relevance
between a token and a passage~$p$.
The intuition behind this design is that the importance of a token can be judged from those documents that are (potentially) highly relevant to it; that is, if a word is representative
of its relevant documents, it is likely to be an important keyword. 

\begin{figure}
\centering
\includegraphics[width=\columnwidth]{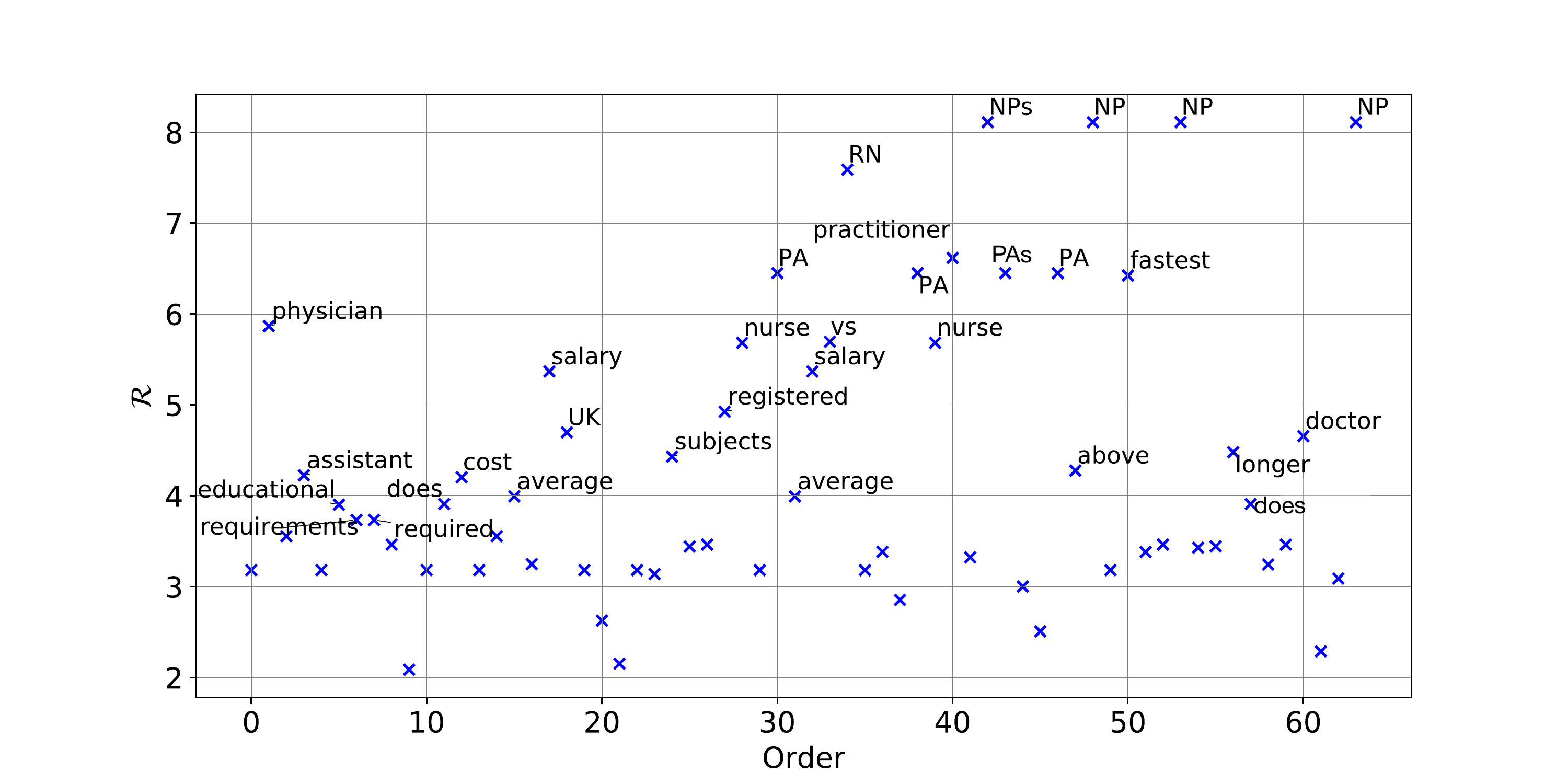}
\caption{Keyword importance distribution for TREC CAsT 2019 Training Topic 1. Each point represents a word from the utterances in Table~\ref{tb:CAsT_example}. The $x$-axis and the $y$-axis are the words' turn order and BM25 retrieval scores, respectively.}
\label{fig:KE}
\end{figure}

Figure~\ref{fig:KE} illustrates the importance ($\mathcal{R}$) of each word in the utterances (listed in their order) in Table~\ref{tb:CAsT_example} using BM25 to compute $\mathcal{F}_{KE}(\cdot,\cdot)$.
As shown in the figure, topic words (e.g., ``physician'', ``PA'', ``RN'', ``nurse'') usually score above 4.5 to 5 and the less~important words have scores around 3.5 to 4.
A similar phenomenon is also shown in our sensitivity analysis in Section~\ref{sec:sens}

\subsubsection{Query Performance Predictor ($QPP$)}
\begin{table}[t!]
	\caption{QPP measurements of conversational utterances from TREC CAsT 2019 Training Topic 1.}
	\label{tb:qpp}
	\centering
	\small
    \begin{tabular}{clr}
	\toprule
   Turn ($i$) & Conversation utterances ($u_i$)  & $\mathcal{A}_i$\\
    \midrule
    1&	What is a physician's assistant? & 11.86\\
    2&	What are the educational requirements required to become one? & 15.01\\
    3&	What does it cost? & 9.56\\
    4&	What's the average starting salary in the UK? & 13.77\\
    5&	What about in the US? & 7.21\\
  
    6&	What school subjects are needed to become a registered nurse? & 19.34\\
  ...&\\
	\bottomrule
	\end{tabular}
\end{table}

Given an utterance $u_i$ and a passage collection $\mathcal{C}$, QPP measures the
utterance's ambiguity.
The literature demonstrates that the degree of query ambiguity is closely
related to its ambiguity with respect to the collection of documents being
searched~\cite{clarity, nqc, wig, qpp2019, qpp2012}; thus, many metrics evaluate
query ambiguity by analyzing retrieval scores.
As we here are focused on providing an effective query expansion
strategy for CQR rather than calculating the most accurate QPP, we keep the
measurement of utterance ambiguity as simple as possible. 
Following the KE, we measure utterance ambiguity for $u_i$ as
\begin{equation*}
\mathcal{A}_i=\text{QPP}\left(u_i, \mathcal{C}\right)=\max\limits_{p\in \mathcal{C}} \left\{ \mathcal{F}_{\text{QPP}}\left(u_i,p\right) \right\},
\end{equation*}
where $\mathcal{A}_i$ stands for the degree of utterance ambiguity and
$\mathcal{F}_{\text{QPP}(\cdot)}$
estimates the relevance score
between a passage and an utterance.
In our experiments, we set $\mathcal{F}_{\text{QPP}(\cdot)}$ and $\mathcal{F}_{\text{KE}}(\cdot)$ to be the BM25 function.
Note that the higher the $\mathcal{A}_i$ score, the more precise the
utterance~$u_i$.

Table~\ref{tb:qpp} illustrates the intuition behind our QPP measurement by showing $\mathcal{A}_i$ for each conversational utterance.
It is clear that turns~3 and 5 have relatively low scores.
This indicates that they fall into the most ambiguous category for IR systems and should be expanded using both topic and subtopic keywords.
The measurement matches our observation of query ambiguity in Observation~\#2, and also indicates that the lower the retrieval score is, the more ambiguous the query is,  similar to the finding in~\citet{Tomlinson04}.

\subsubsection{Putting it all together}
Algorithm~\ref{alg:hqe} details the complete HQE method, $g_{\phi}(\{u_i, u_{<i}\}, \mathcal{C})$: keyword
extraction (lines 3--8), query performance prediction (line 10), and query
expansion (lines 11--13).
Note that $\mathcal{R}_{\text{topic}}$, $\mathcal{R}_{\text{sub}}$ (where
$\mathcal{R}_{\text{topic}}>\mathcal{R}_{\text{sub}}$), $\eta$, and
$\mathcal{M}$ are hyperparameters.
Specifically, for each utterance $u_i$ in a session $s \in S$ and a given
passage collection $\mathcal{C}$, HQE first extracts topic (subtopic) keywords  from
$u_i$ if $\mathcal{R}_j^k>\mathcal{R}_{\text{topic}}$ ($\mathcal{R}_j^k>\mathcal{R}_{\text{sub}}$), and collects them in the keyword set $W_{\rm topic}$ ($W_{\rm sub}$). 
Then, QPP measures the ambiguity of all~$u_{i}$ for $i>1$.
Here $\eta$ is the threshold to judge whether an utterance falls into the 
ambiguous category. 
For all $u_i$ except the first utterance $u_1$, HQE first rewrites $u_i$ by
concatenating $u_i$ with the topic keyword set $W_{\rm topic}$ collected from $u_i$ and $u_{<i}$.  
Moreover, if $u_{i}$ is ambiguous (i.e., $\mathcal{A}_i<\eta$), HQE further
adds the subtopic keywords from the previous $\mathcal{M}$ turns and turn~$i$.  
We thus assume that the first utterance in a session is clear enough and
that following utterances belong to the second or the ambiguous category.  
Note that we concatenate $W_{\rm sub}$ derived from the previous $\mathcal{M}$
turns, as subtopic keywords last a few turns (see Observation~\#1 for the characteristics of subtopics in conversation). 
Also note that $W_{\rm sub}$ includes the topic keywords in
$W_{\rm topic}$, which ensures that topic keywords gain higher term weights
than subtopic keywords in rewritten utterances.

    \begin{algorithm}[t!]
	\small
	\caption{Historical Query Expansion}
	\label{alg:hqe}
	\KwIn{ ${u_i, u_{<i}}, \mathcal{C}$}
	\KwOut{$\bar{u}_i$}
    $\bar{u}_i\leftarrow ()$;
	$W_{\text{topic}} \leftarrow \{\}$; $W_{\text{sub}} \leftarrow \{\}$\\
	\For{$j=1$ to $i$}{
		\For{$k=1$ to $n(u_j)$}{
    		$\mathcal{R}^k_j=\text{KE}\left(t_j^k, \mathcal{C}\right)$\\
    		\If{$\mathcal{R}_j^k>\mathcal{R}_{{\rm topic}}$}{
        		$W_{\text{topic}}$.insert$\left(t_j^k\right)$\\
    		}
    		\If{$(\mathcal{R}_j^k>\mathcal{R}_{{\rm sub}})$  {\bf and} $(j \geq i-\mathcal{M})$}{
        		$W_{\text{sub}}$.insert$\left(t_j^k\right)$\\
    		}
		}
	}
	\If{$i>1$}{
		$\mathcal{A}_i=\text{QPP}(u_i, \mathcal{C})$\\
		$\bar{u}_i$.insert$\left(t\right)$ for all $t \in W_{\rm topic}$\\
		\If{$\mathcal{A}_i<\eta$}{
    		$\bar{u}_i$.insert$\left(t\right)$ for all $t \in W_{\rm sub}$\\
		}
	}
    $\bar{u}_i$.append($u_i$)
	
	\BlankLine
	\KwRet{$\bar{u}_i$}\\
	\end{algorithm}

\begin{figure}[ht!]
\centering
\includegraphics[width=.35\columnwidth]{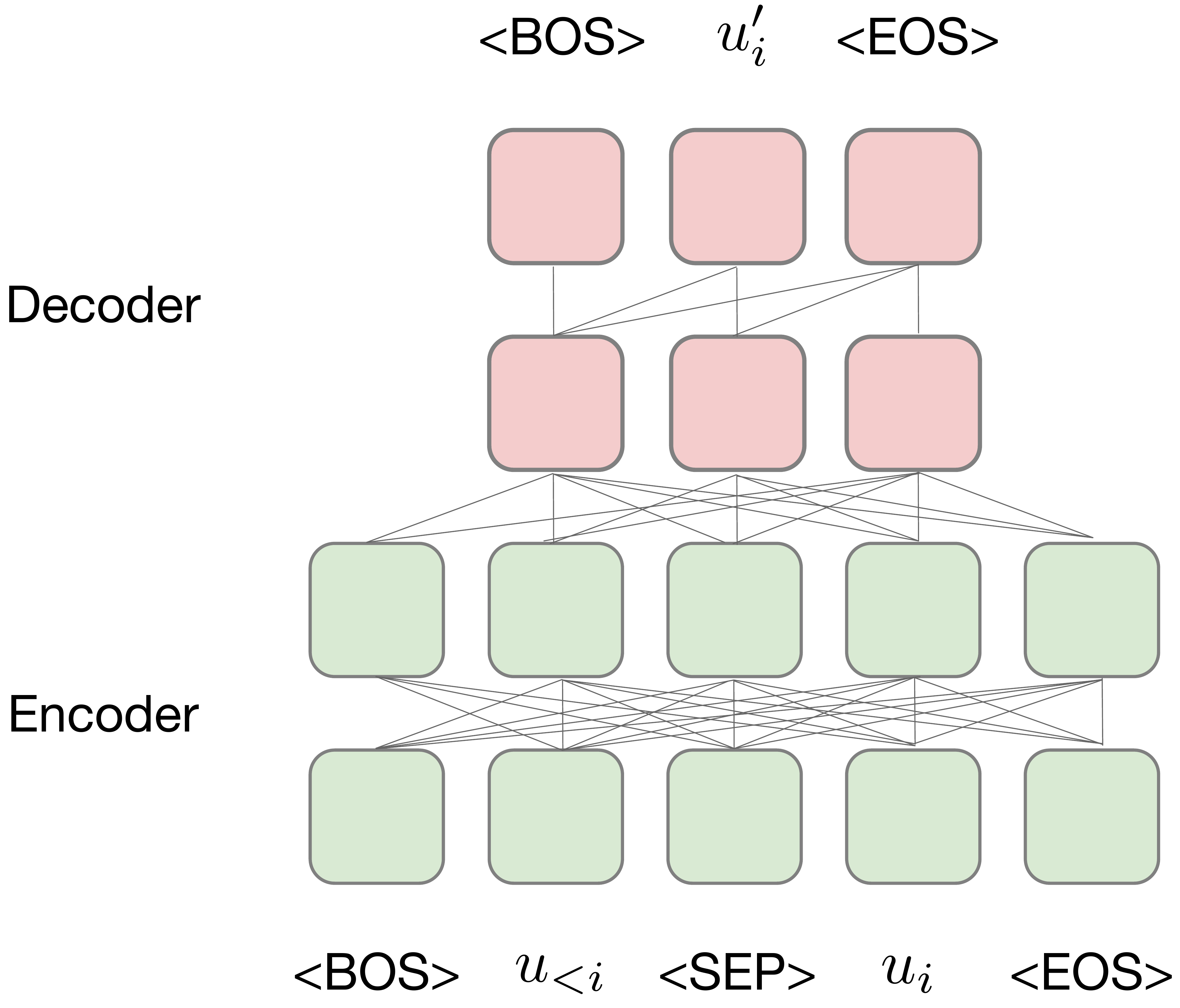}
\caption{
Transformer~\cite{transformer}, the encoder-decoder architecture we use for the CQR task; in this work, we specifically use T5.
Taken from~\citet{t5}, in this diagram, blocks are used to represent elements in a sequence and lines to represent attention visibility.
<BOS>, <SEP>, and <EOS> stand for special tokens placed in the input sequence to indicate begin of sequence, separation, and end of sequence, respectively.
Whereas attention in the encoder part is bidirectional, the decoder part follows a one-directional pattern called ``causal masking''.
We choose this architecture to encode all conversational context $(u_{<i}, u_{i})$ bidirectionally and let the decoder generate $u_{i}'$ token by token, conditioned on all the contexts.
}
\label{fig:encoder_decoder}
\end{figure}

\subsection{Neural Transfer Reformulation}\label{sec:T5}

Communicating information through dialogues of concise representations in texts
is natural for humans.
In these dialogues, humans often use colloquially expressed and contextually dependent queries in a sequence of utterances, as shown in Table~\ref{tb:CAsT_example}.
These concise queries, distinguished by natural language properties---coreferences,
omissions, and abbreviations---are easily resolved by humans but
understood by traditional ad-hoc retrieval systems only with great difficulty.
Therefore, before passing these queries to downstream components that can
only digest a single and clear query at a time (e.g., traditional ad-hoc IR
systems), these natural language properties must first be resolved.
To address this, previous studies~\cite{cqu, CANARD,
lin2020conversational, vakulenko2020question} adopt data-driven models to
mimic patterns of how humans rewrite queries given conversational contexts.
Following this line of research, we propose reformulating a raw
utterance $u_i$ into a coreference-and-omission-free natural language query
$q^{\rm NL}_{i}$ using our proposed method, called Neural Transfer Reformulation (NTR).
Specifically, we fine-tune a pretrained seq2seq language model as a
conversational LM, which we then leverage to learn to
mimic and transfer patterns of how people rewrite questions in a
conversational query rewriting task from another large-scale annotated dataset.
Compared to the aforementioned HQE approach, reformulated queries of  NTR generally retain their linguistic structures.

\subsubsection{Architecture}

Transformer, a popular and highly successful sequence-to-sequence model, was originally introduced with
an encoder-decoder architecture using an attention mechanism by~\citet{transformer}.
Based on this architecture, various studies on transfer learning for NLP
have proposed variants that yield state-of-the-art
effectiveness on a range of NLP benchmarks by pretraining using
different language modeling objectives~\cite{BERT, unilm, gpt2}.
Instead of providing a comprehensive review of various Transformer
architectures, we refer interested readers to the original
paper~\cite{transformer} and a follow-up work:\ the Text-to-Text Transfer
Transformer (T5)~\cite{t5}, which provides a detailed
introduction and explores the limits of transfer learning with 
encoder-decoder Transformer architectures on NLP tasks such as natural language
understanding, reading comprehension, translation, and summarization.

An illustrative example of CQR with a Transformer encoder-decoder
architecture is shown in Figure~\ref{fig:encoder_decoder}.
Given a raw query $u_{i}$ and its variable-length conversational context
$u_{<i}$, our objective is to train the Transformer model to unpack and
generate an explicit standalone query $u_{i}^{'}$ that entails
equivalent answers to $u_{i}$~\cite{CANARD}.
Here we use the encoder to consume $u_{i}$ and its context $u_{<i}$, and use the decoder to generate tokens of $u_{i}^{'}$.
To obtain the explicit queries, we employ teacher forcing and 
cross-entropy loss, which are standard techniques for training language
generation models, to train the Transformer to predict every token in the
output sequence given all of the preceding tokens.

\subsubsection{Training strategy}
So far we consider the setting of performing the CQR task with the encoder-decoder architecture.
While this approach is straightforward, various alternatives for training the model on CQR datasets have been proposed~\cite{cqu, CANARD, vakulenko2020question, lin2020conversational}.
In this section, we illustrate how we transfer the knowledge from
a specific CQR dataset to reformulate raw queries in TREC CAsT 2019 via the
encoder-decoder architecture.

First, we require three ingredients to use NTR to construct the 
function $g_{{\theta}}$ introduced in Section~\ref{sec:multi-stage}:
(a) a large-scale, high-quality dataset of human-generated queries $q^{\rm NL}$ with source utterances and contexts;
(b) an architecture to map an utterance and its conversational context into $q^{\rm NL}$; and 
(c) a dataset with enough diversity to cover open-domain exploratory information needs selected from the session sets of our interest $S^I$.
Fortunately, open-domain QA research has produced QuAC~\cite{quac}, a diverse, large-scale dataset that contains conversational natural language questions of exploratory information needs, as well as CANARD~\cite{CANARD}, a derived conversational question-in-context rewriting dataset with human-generated questions for QuAC questions.

Second, as with CANARD and text summarization studies~\cite{CANARD,
transformer_sum}, we choose a sequence-to-sequence~\cite{seq2seq,
cho-etal-2014-learning} (seq2seq) architecture to map variable-length
conversational contexts $u_{<i}$ and $u_i$ into an explicit standalone query $q^{\rm
NL}_{i}$ (or $u_{i}'$ in Figure~\ref{fig:encoder_decoder}).
Without loss of generality, instead of using $\theta$ to parameterize the function $g_{\theta}$ that reformulates conversational queries optimized for an IR system in Eq.~(\ref{eq:multi_stage}), 
we define a function parameterized by ${\bar{\theta}}$, taking input
tokens $(x_1,\dots, x_n)$ of length $n$ and generating output tokens $( y_1,\dots, y_m
)$ of length $m$ as
\begin{equation*}
    \text{seq2seq}\left( (x_1,\dots, x_n), \bar{\theta} \right) = ( y_1,\dots, y_m )
\end{equation*}
for this neural query reformulation task.
A proxy for obtaining a particular set of parameters $\hat{\theta}$ for this task under the configuration of a function $\text{seq2seq}( {\cdot, \hat{\theta}} )$ and a dataset from CANARD instead of CAsT is then
\begin{equation}
\label{eq:ntr}
     \hat{\theta} = \mathop{\arg\max_{\bar{\theta}}} \prod_{i} P\left( \bar{u}_{i}'\,|\,\text{seq2seq}(\left[\bar{u}_{<i}\,\Vert\,\bar{u}_i\right], \bar{\theta}) \right),
\end{equation}
where $[\bar{u}_{<i}\Vert \bar{u}_i]$ stands for the concatenation of a
conversational context and an utterance of the $i$-th turn defined in the CANARD dataset with the separation token ``|||'' that indicates the boundary of utterances from different conversation turns.
Note that here we use $\bar{\theta}$ to indicate a set of parameters initialized before optimizing for the CQR task.
The symbols $\bar{u}_{<i}$ and $\bar{u}_i$ represent utterances from the CANARD dataset for a CQR task instead of those in the CAsT dataset for a ConvPR task.
To rewrite a raw utterance $u_{i}$ at the $i$-th turn into an explicit utterance $u_{i}'$, we construct training triples $(\bar{u}_{<i}, \bar{u}_i, \bar{u}_{i}')$ by pairing each explicit utterance $\bar{u}_{i}'$ with its ambiguous counterpart $\bar{u}_i$ and its context $\bar{u}_{<i}$ from the CANARD dataset.\footnote{
Here we only consider ``raw'' queries in previous turns as historical context when constructing training triples, not the explicit queries annotated by humans.
That is to say, the explicit queries here serve only as the targets to be generated instead of part of the context.}
We learn the optimized parameter set $\hat{\theta}$ using standard techniques for training a neural seq2seq model:\ cross-entropy loss, teacher forcing, and stochastic gradient
descent.

Finally, to rewrite queries for a ConvPR task after training the model on the CQR dataset, we adopt parameter and
network architecture sharing as a simple strategy for transfer learning.
Thus, after training on CANARD, we directly use the seq2seq model with its optimized parameter set $\hat{\theta}$ to form our model $g_{{\theta}}(\cdot)$
(i.e., $\theta=\hat{\theta}$) and directly use the model $g_{{\theta}}(\cdot)$
to reformulate $q_i$ from an information set $\{ u_i, u_{<i} \}$ from the CAsT dataset.
To this end, we introduce a general scheme for training a seq2seq (encoder-decoder) model on the human-annotated CQR dataset and transferring it to a ConvPR dataset.
Although this seq2seq model can be trained from scratch, i.e., randomly initialized $\bar{\theta}$ in Eq.~(\ref{eq:ntr}), as is done in other work~\cite{cqu,
CANARD}, a previous study has shown the effectiveness of leveraging Transformers pretrained with different language modeling objectives~\cite{lin2020conversational} before training on the CQR task itself.
Furthermore, the end-to-end integration and evaluation of applying the queries generated by the seq2seq model on the ConvPR task is addressed only for decoder-only architectures~\cite{vakulenko2020question}.
In the following sections, we evaluate the effectiveness of integrating queries generated from our proposed NTR module, which is pretrained with a masked language modeling objective introduced by~\citet{t5} and then fine-tuned on the CQR dataset with standard techniques used to train a neural seq2seq model (described above).

\subsection{Combining Query Variations via Rank Fusion}\label{sec:fusion}

Combining query variations that represent users' information needs in different ways is a successful approach to improve retrieval effectiveness~\cite{belkin1993}.
However, to our best knowledge, there is no such discussion for the ConvPR task.
Thus, we introduce two types of CQR methods, both of which aim to resolve users' information needs embedded in the conversation history but from different points of view.
To incorporate the idea of combined query variations into our multi-stage ConvPR pipeline, we adopt a previously proposed fusion technique~\cite{rrf} to combine query variations from multiple CQR systems.
Specifically, given a passage list $L_{j}$ in which the passages are ranked by relevance scores from system $j$ given query $q$, we leverage the reciprocal rank scoring function proposed by~\citet{rrf} to aggregate the scores from different systems:
\begin{equation}
\label{eq:fuse}
    \text{RRF}(p) = \sum_{j} \frac{1}{k + \text{rank}(p, L_{j})},
\end{equation}
where $k$ is a constant and $\text{rank}(p,L_j)$ denotes the rank of passage~$p$ in $L_j$.\footnote{We set $k = 60$, following~\citet{rrf}.}

We fuse the query variations by replacing query $q_i$ reformulated from utterances $\{u_i, u_{<i}\}$ using the $j$-th query reformulation model $P_{j}(q_i|\{u_i, u_{<i}\})$, and obtain the ranked passage list $L_{j}$ using the relevance prediction models, which compute $P(R = 1|q_i, p)$.
Note that in the proposed multi-stage ConvPR pipeline, there are two types of relevance prediction models:\ BM25 for the retriever and BERT for the re-ranker, as shown in the two panels of Figure~\ref{fig:configuration}.
Hence, there are two possible configurations to combine the query variations via Eq.~(\ref{eq:fuse}):
\begin{enumerate}
    \item Late fusion that performs fusion after the entire retrieval and re-ranking process, shown in Figure~\ref{fig:configuration}(a);
    \item Early fusion that performs fusion at the retrieval stage, shown in Figure~\ref{fig:configuration}(b).
\end{enumerate}
In this paper, we study both configurations to analyze the queries reformulated from different CQR models, to understand their effects on multi-stage ConvPR pipelines.

\section{Experiment Setup}

We empirically evaluate the effectiveness of our proposed conversational multi-stage retrieval system.
Specifically, we apply the system to the passage retrieval task defined by TREC CAsT 2019 and also compare different types of CQR methods in our multi-stage pipeline.

\subsection{Datasets}
\textbf{CANARD.}
\label{sec:canard}
CANARD~\cite{CANARD} is a conversational query rewriting dataset manually built on QuAC~\cite{quac}, a conversational question answering dataset.
CANARD contains 40,527 query pairs, each of which consists of an original query from a dialogue session and a reformulated query free from coreference and omissions produced by human annotators according to its context in the dialogue.
The query pairs are split into training, development, and test sets (with 31.5k, 3.4k, and 5.5k pairs, respectively).
In our experiments, we use only the training set for NTR model training.

\smallskip \noindent
\textbf{CAsT.}
\label{sec:cast}
We conduct experiments on the dataset provided by the TREC~2019 Conversational
Assistant Track (CAsT)~\cite{cast}, a new task for research on conversational search.
The dataset consists of training and evaluation sets with 30 and 50
sessions, respectively, covering a wide range of open-domain topics. 
Each session contains approximately 10 turns, and each turn includes a query and a list of relevant passages from the corpus, which is comprised of passages from the MS MARCO Passage Ranking collection and the TREC CAR paragraph collection (v2.0).
Near-duplicate passages in the corpus are eliminated with the TREC CAsT
tools,\footnote{\url{https://github.com/gla-ial/trec-cast-tools}} yielding a 
total of approximately 40 million candidate passages. 
As shown in Table~\ref{tb:statistics}, $13$ of the $30$ training set sessions have relevance judgments and $20$ of the $50$ evaluation set sessions have relevance judgments for final evaluation. 

\begin{table}[t]
	\caption{TREC CAsT 2019 dataset statistics.}
	\label{tb:statistics}
	\centering
	\small
    \begin{tabular}{lrr}
	\toprule
	& Training\footnotemark & Evaluation\\
	\midrule
	    Sessions (topics) &13 & 20    \\
	    Turns &108 & 173 \\
	    Assessments &2,399 & 29,571 \\
	    \midrule
	    Fails to meet (0)   &1,759 & 21,451 \\
	    Slightly meet (1)   &329   & 2,889 \\
	    Moderately meet (2) &311   & 2,157 \\
	    Highly meet (3)     &0     & 1,456 \\
	    Fully meet (4)      &0     & 1,618 \\
	\bottomrule
	\end{tabular}
\end{table}
 \footnotetext{Note that training judgments are only graded on a three-point scale (2: very relevant,~1: relevant, and 0: not relevant).}

\subsection{Baseline Query Reformulation Methods}
\label{sec:qr_setting}

\smallskip \noindent
\textbf{Best CAsT entry.} 
This baseline is one of our submissions to TREC CAsT 2019, which uses an earlier version of the proposed HQE method in a two-stage ConvPR system.
This method was the best automatic run of 41 submissions from 21 teams.

\smallskip \noindent
\textbf{Raw query.} 
A simple baseline that uses the original queries without any query reformulation. 

\smallskip \noindent
\textbf{Concat.}
Another simple baseline that concatenates each query with the queries in its previous
$\mathcal{M}$ turns, where $\mathcal{M}$ is a hyperparameter. 
A variant of this method is to filter out certain types of words from the queries
in the previous $\mathcal{M}$ turns before concatenation.
Specifically, we discard words with POS tags other than adjective and noun, using spaCy as the POS tagger; this condition is denoted as Concat (+POS).\footnote{\url{https://github.com/explosion/spaCy}}
This filter is also applied to the proposed HQE method, denoted as HQE (+POS).  

\smallskip \noindent
\textbf{Manual.}
The TREC CAsT organizers manually rewrote the originally ambiguous queries according to the conversation context.\footnote{\url{https://github.com/daltonj/treccastweb}}
As the rewritten queries contain all the context information
 packaged into a single query, in the experiments we consider this condition to be the empirical upper bound on human effectiveness.

\subsection{Evaluation and Settings}

\smallskip \noindent
\textbf{Information retrieval model settings.} 
As mentioned in Section~\ref{sec:multi-stage}, we implement a two-stage
information retrieval pipeline with BM25 retrieval (first stage) and BERT
re-ranking (second stage). 
The parameters for the BM25 model are $k_1=0.82$ and $b=0.68$ and the
number of retrieved passages is set to $1000$.
We use the Anserini toolkit~\cite{anserini} for indexing and retrieval.
The fine-tuned BERT-large (uncased) model for the second stage re-ranking is provided by~\citet{marco_BERT}; we use the model fine-tuned on the MS MARCO passage dataset.

\smallskip \noindent
\textbf{Query reformulation model settings.}
The hyperparameters are selected by grid search on the TREC CAsT 2019  training set (see Section~\ref{sec:sens} for details).
The neural models (i.e., LSTM and T5) are directly applied to rewrite queries with beam search decoding after training on the CANARD dataset.
The detailed settings are as follows:

\begin{itemize}
\item \textbf{LSTM (+Atten.)}:\ We adopt the bi-LSTM seq2seq model with attention, copy mechanism, and the same hyperparameter settings proposed in~\citet{CANARD}.\footnote{{\url{https://github.com/aagohary/canard}}}
\item \textbf{QuReTec}~\cite{quretec}:\ A neural query resolution method based on the BERT-large model. In our experiments, we directly use the queries provided by the authors.\footnote{\url{https://github.com/nickvosk/sigir2020-query-resolution}}
\item \textbf{T5}~\cite{t5}:\ We use the T5-base model\footnote{\url{https://console.cloud.google.com/storage/browser/t5-data/pretrained_models/base}} and its pretrained weights as the initialization and then fine-tune it with a constant learning rate of 1e-3 and a batch size of 256 for 4K iterations.
The maximum input and output token lengths are set to 512 and 64, respectively.
During inference, we set beam width to 10 for beam search.   
\end{itemize}

\smallskip \noindent
\textbf{Evaluation.} 
For both stages, the results are evaluated in terms of the overall ranking metric, mean average precision (MAP) at depth~$1000$, and the early-precision metrics
NDCG@3 and NDCG@1.
Note that NDCG@3 is the main metric used in CAsT.
In addition, we report recall at depth $1000$ (R@1000) for first-stage retrieval.
The evaluation is conducted using the \texttt{trec\_eval} tool,\footnote{\url{https://github.com/usnistgov/trec_eval}} and significance tests are performed by comparing the eight automatic QR methods (Raw/Concat/HQE/NTR + variants) with paired $t$-tests ($p$-value $< 0.05$) for each metric. 
We also provide Win/Tie/Loss counts based on R@1000 and MAP to show
the number of queries whose effectiveness improved/unchanged/deteriorated
compared to manual query reformulation.

\section{Results}\label{sec:fullrank}

In this section, we first examine the effectiveness of the proposed HQE
and NTR methods on the TREC CAsT 2019 dataset; we also provide results and analyses in terms of turn depth.
Second, we study the impact of different query reformulation methods on passage re-ranking and conduct a further fusion analysis of query variations.

\subsection{Main Results}

\begin{table*}[t!]
	\caption{Experimental results on the TREC CAsT 2019 evaluation set. Win/Tie/Loss denotes the number of queries whose effectiveness improved/unchanged/deteriorated compared to manual query reformulation. The best results among automatic query reformulation methods are in bold; superscripts denote significant improvements ($p$-value $< 0.05$) over methods indexed on the left.}
	\label{tb:full_ranking}
	\centering
    \small
    \resizebox{\textwidth}{!}{
    \begin{tabular}{rcllclclllcll}
    \toprule
        & && \multicolumn{6}{c}{BM25}& \multicolumn{4}{c}{Full ranking (BM25+BERT re-ranking)}\\
        \cmidrule(lr){4-9}\cmidrule(lr){10-13}
	    \multicolumn{3}{c}{Query reformulation}  & R@1000 &W/T/L & MAP &W/T/L & NDCG@3 & NDCG@1 & MAP &W/T/L & NDCG@3 & NDCG@1 \\
	\cmidrule(lr){1-3}\cmidrule(lr){4-5}\cmidrule(lr){6-7} \cmidrule(lr){8-9}\cmidrule(lr){10-11}\cmidrule(lr){12-13}
    \multicolumn{3}{c}{Best CAsT entry}& - &-  &- &-  &- &- &0.267 &-  &0.436 &- \\
    \midrule
	    \multicolumn{3}{c}{Manual}&0.801 &- &0.258 &-  &0.309  &0.302 &0.394 &- &0.579 &0.597 \\
        \cdashline{1-13}
	  1& \multicolumn{2}{c}{Raw query}&0.418 &4/58/111 &0.107 &10/21/142  &0.131  &0.136 &0.175 &5/55/113 &0.266 &0.262 \\
		\cdashline{1-13}
		2&\multirow{2}{*}{Concat} &Raw & 0.492 &12/27/134 &0.094 &10/21/142 & 0.179  &0.178 &0.180 &8/21/144 &0.331 &0.348 \\
	3&	&+POS & 0.680 &29/50/94 &0.160 &38/32/103 &0.236  &0.269 &0.267 &28/21/124 &0.423 &0.458 \\
        \cdashline{1-13}
		4&\multirow{2}{*}{HQE}&Raw &0.676 &42/44/87 &0.197 &44/22/107  & 0.257  &0.273 &0.273 &25/21/127 &0.435 &0.453  \\
		5&&+POS &0.730 &44/61/68 &0.211 &47/27/99  &0.259 &0.261 &0.304 &29/21/123 &0.481 &0.485 \\
		\cdashline{1-13}
		6&\multirow{3}{*}{NTR}     & LSTM (+Atten.) &0.527 &13/57/103&0.136 &12/49/112  &0.178  &0.163 &0.230 &24/36/113 &0.359 &0.360 \\
   7& &QuReTeC&\textbf{0.771}$^{1-6}$ &23/107/43 & \textbf{0.236}$^{1-6}$ &29/88/56 & \textbf{0.296}$^{1-6}$ &0.291&0.350 &55/10/108  &0.503 &0.526 \\
		8&&T5 & 0.744 &11/125/37 &0.223 &11/111/51 &0.295 &\textbf{0.293}$^{1-2,6}$&\textbf{0.359}$^{1-6}$ &63/20/90  &\textbf{0.556}$^{1-7}$ &\textbf{0.596}$^{1-7}$ \\
        
        \midrule
\multirow{2}{*}{9}&\multirow{2}{*}{Fusion} &HQE (+POS) &\multirow{2}{*}{0.804$^{1-8}$} &\multirow{2}{*}{58/68/47}&\multirow{2}{*}{0.255$^{1-8}$} &\multirow{2}{*}{73/24/76}  &\multirow{2}{*}{0.325$^{1-6,8}$}  &\multirow{2}{*}{0.337$^{1-6,8}$} &\multirow{2}{*}{0.375$^{1-8}$} &\multirow{2}{*}{77/24/72} &\multirow{2}{*}{0.565$^{1-7}$} &\multirow{2}{*}{0.598$^{1-7}$} \\
&&NTR (T5)& \\
 \bottomrule
	\end{tabular}
	}
\end{table*}

\smallskip \noindent
\textbf{Full ranking.}
Table~\ref{tb:full_ranking} (``Full ranking'' columns on right)
shows the end-to-end results with the two-stage ConvPR approach on the
TREC CAsT 2019 evaluation set.
The listed metrics are from the re-ranked results based on the
corresponding 1000 retrieved passages obtained in the first stage using BM25, the metrics of which can be found in the same row.  
We note that all the query reformulation methods outperform the
baseline with raw queries; the naive Concat method serves as a competitive
baseline.
Our proposed HQE and NTR methods beat the best entry in TREC CAsT 2019. 
In particular, HQE (+POS) outperforms the best CAsT entry, which represents our work based on an earlier version of HQE~\cite{hqae}, in terms of MAP and NDCG@3 by $13\%$ and $10\%$, respectively, whereas NTR (T5) significantly surpasses the best entry by $34\%$ in MAP and $30\%$ in NDCG@3.

Comparing the results of Concat with and without the POS filter suggests that using only adjectives and nouns helps accurately identify important keywords from historical queries.
Although the POS filter further improves the effectiveness of HQE, the method without such filtering still yields competitive effectiveness, thus validating the design of the proposed keyword extraction module.
For the neural models, the LSTM trained from scratch performs poorly;
in contrast, fine-tuned NTR (QuReTec/T5) delivers far better results, illustrating that the pretrained weights provide a satisfactory initialization for neural query reformulation models.
Among all CQR methods, NTR (T5) yields the best ranking quality and significantly outperforms other CQR methods in terms of NDGC@3 and NDCG@1.

\smallskip \noindent
\textbf{First-stage retrieval with BM25.}
The effectiveness of the proposed query reformulation methods can also be observed from the results using only the BM25 retriever at the first stage.
As shown in Table~\ref{tb:full_ranking} (``BM25'' columns on the left), the queries reformulated by HQE (+POS) and NTR (T5) outperform the other baselines, except for NTR (QuReTec), leading to average effectiveness improvements in terms of R@1000 of over $70\%$ and MAP of around $20\%$.
However, the Win/Tie/Loss comparison with manual queries shows that the two methods are effective in markedly different ways.
Specifically, fewer than $30\%$ of the queries reformulated by NTR (T5) are worse than their manual counterparts, which is far better than HQE (+POS), which is worse for $40\%$ and $57\%$ of the queries based on R@1000 and MAP, respectively.
On the other hand, HQE (+POS) shows over 40 wins out of 173 queries, whereas only around 10 NTR (T5) rewritten queries surpass the manual queries.
It is a surprising finding that these two methods achieve similar average recall, but in such different ways; we explore this further with a detailed analysis in Section~\ref{sec:discuss}.

\smallskip \noindent
\textbf{Fusion.}
As HQE (+POS) and NTR (T5) represent users' information needs in different ways, we examine fusing the BM25 retrieval lists generated from HQE (+POS) and NTR (T5) with reciprocal rank fusion and further re-rank the fused list using queries generated by NTR (T5).
The effectiveness of this approach, which we call early fusion, is shown in the last row of Table~\ref{tb:full_ranking}.
Observe that the fusion of the two lists in the first stage significantly
outperforms the individual sources and even yields effectiveness comparable to manual queries, leading to more wins than losses in terms of both R@1000 and MAP.
Furthermore, the higher-quality fused list leads to state-of-the-art full ranking results of $0.375$ MAP and $0.565$ NDCG@3.
Additional analyses of the fusion variants are explored in Section~\ref{subsec:fusion}.

\begin{figure*}[htb!]
\begin{subfigure}[t]{.4\columnwidth}
    \includegraphics[width=\columnwidth]{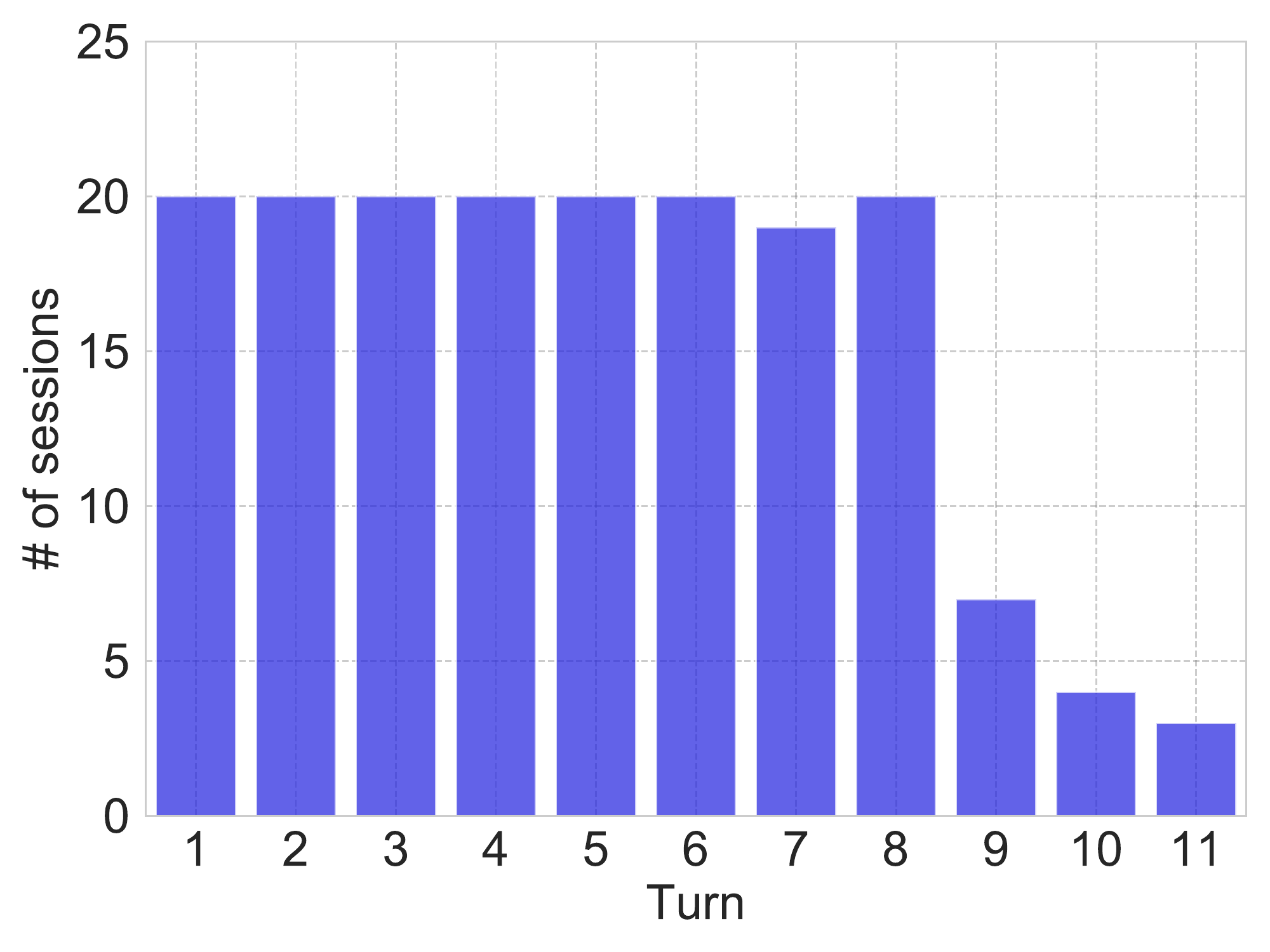}
    \caption{Sessions (per turn)}
\end{subfigure}
\begin{subfigure}[t]{.4\columnwidth}
    \includegraphics[width=\columnwidth]{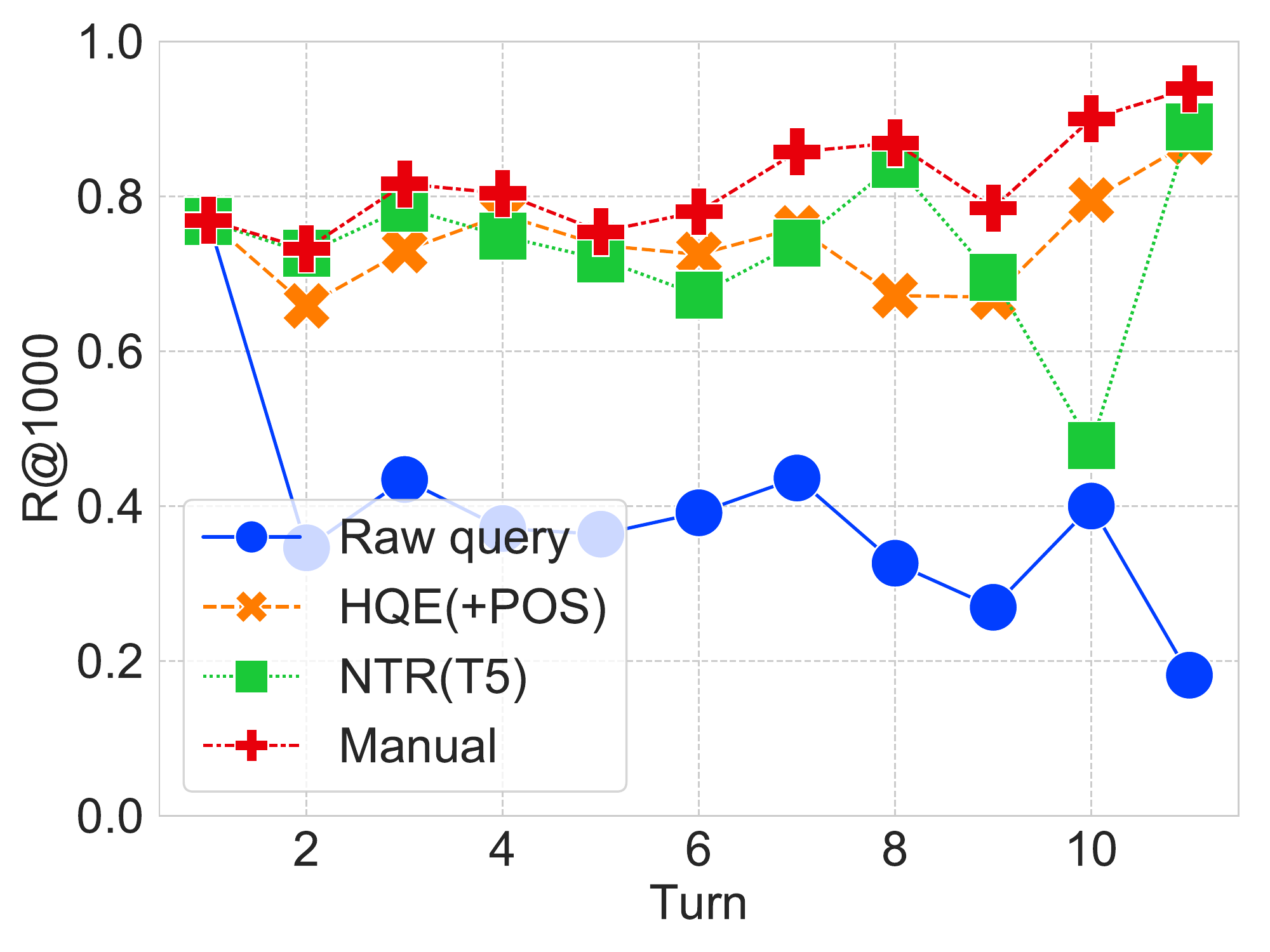}
    \caption{BM25 (R@1000)}
\end{subfigure}
\begin{subfigure}[t]{.4\columnwidth}
    \includegraphics[width=\columnwidth]{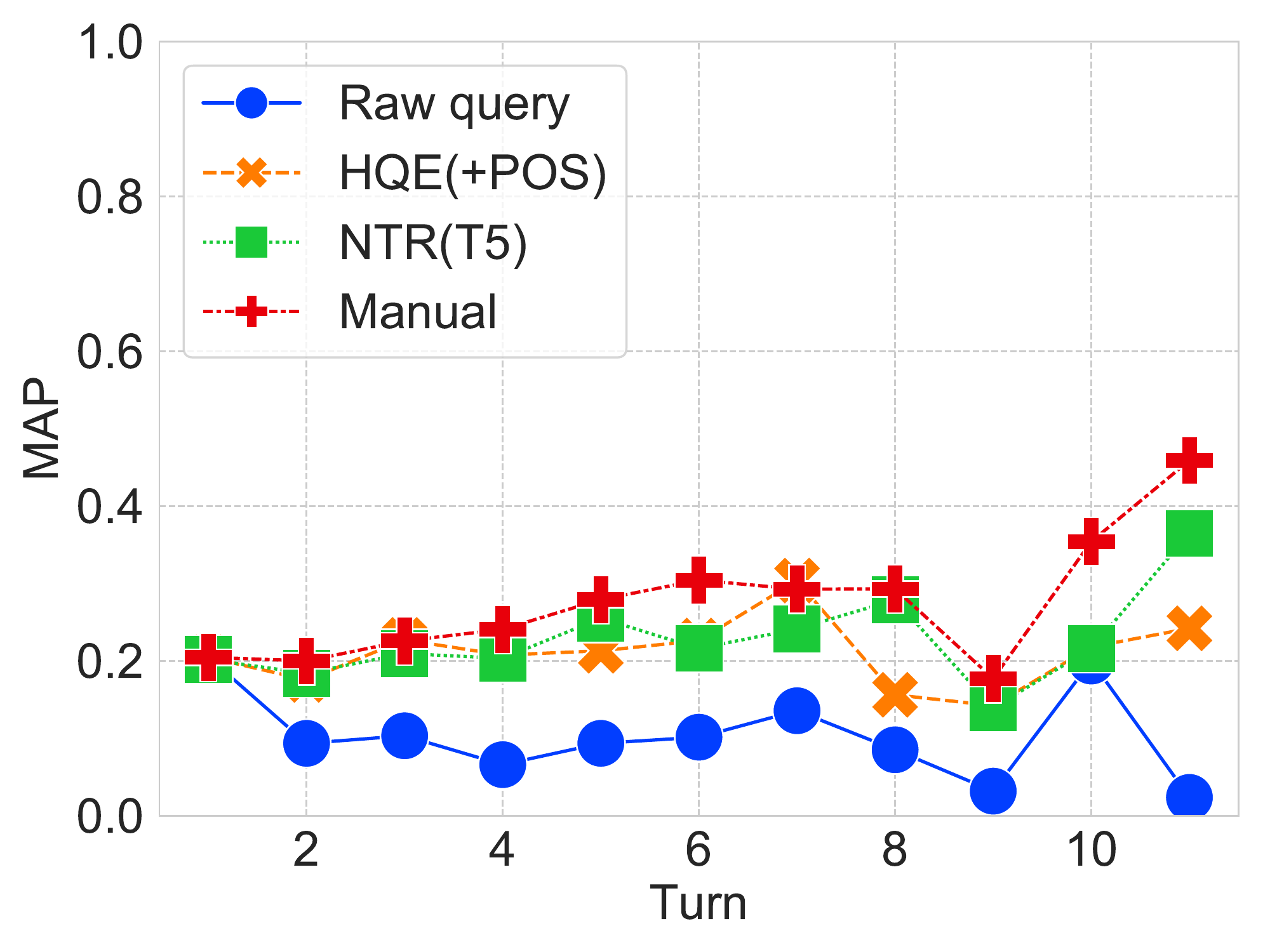}
    \caption{BM25 (MAP)}
\end{subfigure}
\begin{subfigure}[t]{.4\columnwidth}
    \includegraphics[width=\columnwidth]{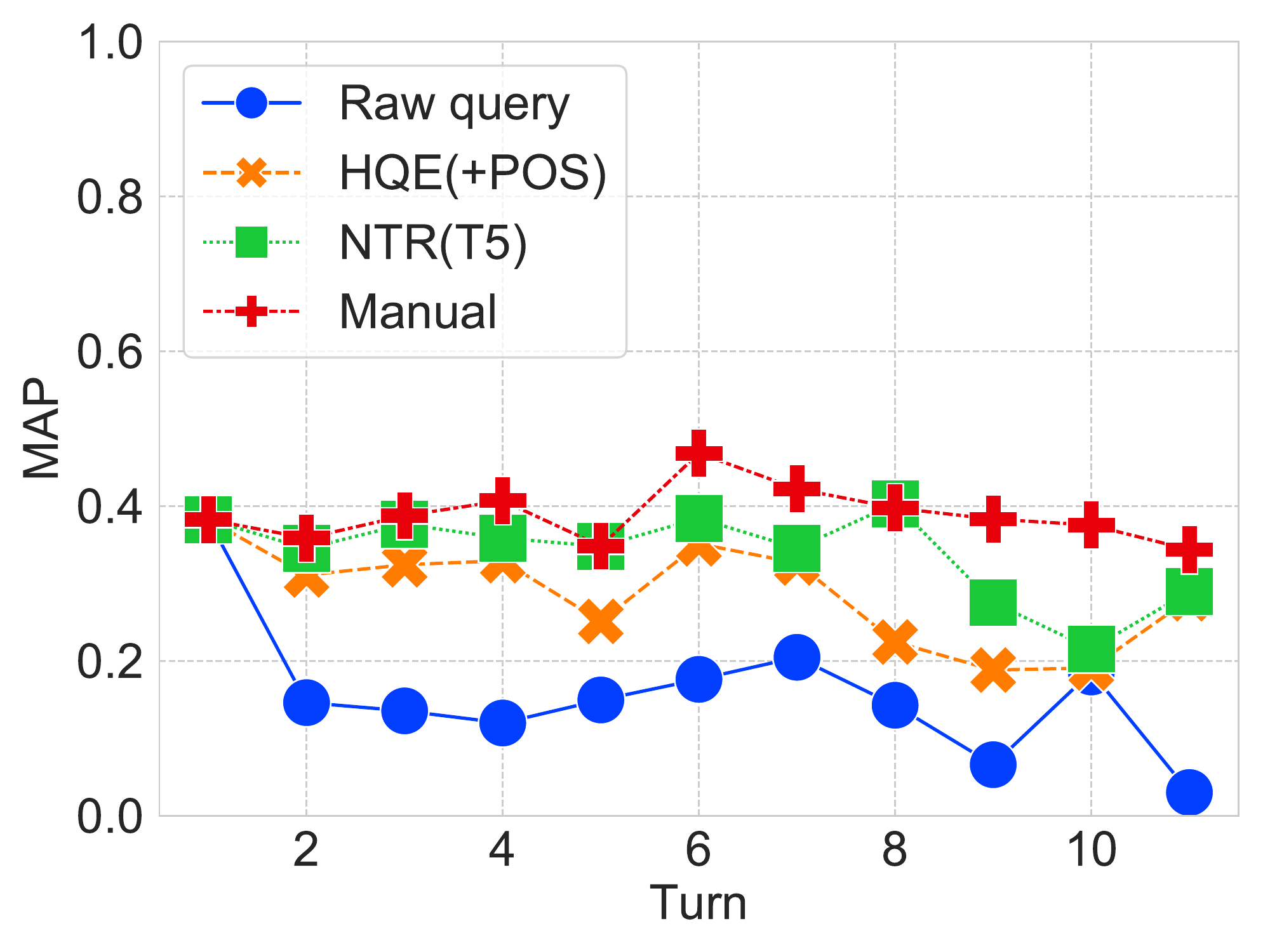}
    \caption{Full ranking (MAP)}
\end{subfigure}
\caption{Retrieval effectiveness comparison by turn depth.}
\label{fig:turn_depth}
\end{figure*}

\smallskip \noindent
\textbf{Results by turn depth.}
Figure~\ref{fig:turn_depth} compares the effectiveness of different reformulated queries in terms of the conversational turn depth.
Not surprisingly, the effectiveness of raw queries (blue lines) degrade  after the first turn:\ conversational queries by nature become ambiguous as a dialogue moves forward.
In contrast, the effectiveness of HQE (+POS) and NTR (T5) remains relatively stable across turns with only slightly worse quality than the manual queries.
In terms of ranking effectiveness, HQE (+POS) sees an obvious drop after
the $7$th turn in both stages, whereas NTR (T5) shows a slight effectiveness
decrease after the $9$th turn, which suggests an advantage of NTR (T5) over
HQE (+POS) in tracking deep conversations.

\smallskip \noindent
\textbf{Summary.}
We provide two strong query reformulation methods for conversational passage retrieval:\ HQE and NTR.
Our experiments demonstrate that the reformulated queries significantly improve the effectiveness of BM25 first-stage retrieval and BERT re-ranking.
Furthermore, the two methods significantly outperform the best TREC CAsT 2019 entry, and NTR (T5) yields the best full ranking accuracy among all CQR methods.
Our analysis also shows that HQE (+POS) and NTR (T5) improve query reformulation from different perspectives and that fusing the two methods leads to state-of-the-art effectiveness for the CAsT full ranking task.
It is also worth mentioning that although NTR (QuReTeC) yields slightly better BM25 retrieval effectiveness than NTR (T5), in the full ranking condition, NTR (T5) still outperforms all the other CQR methods, especially in terms of early precision. 

\subsection{Query Variation Effects on Re-ranking}
\label{sec:reranking}
The retrieval effectiveness of full ranking does not fairly reflect the effects of each query reformulation method on BERT re-ranking, as it is also affected by the number of relevant passages retrieved in the first stage.
Therefore, we conduct another experiment to examine the effects of
re-ranking in isolation.
Specifically, we first retrieve the top 1000 passages using manual queries with BM25 and re-rank the passages using the reformulated queries from different query reformulation methods, and then we feed these queries to the BERT re-ranking module.
In this setting, all the reformulation approaches have the same passage candidates pool for re-ranking, ensuring a fair comparison.

\begin{table}[tb!]
	\caption{Re-ranking passages retrieved with manually rewritten queries. Superscripts denote significant improvements ($p$-value $< 0.05$) over the labeled QR methods.}
	\label{tb:reranking}
	\centering
	\small
    \begin{tabular}{rcllcll}
	\toprule
	    \multicolumn{3}{c}{Query reformulation}  & MAP  &W/T/L  & NDCG@3 & NDCG@1   \\
	\cmidrule(lr){1-3}\cmidrule(lr){4-5}\cmidrule(lr){6-7}
	    \multicolumn{3}{c}{Manual} & 0.394&-& 0.579 & 0.597 \\
        \cdashlinelr{1-7}
	    1&\multicolumn{2}{c}{Raw query} &0.227  &7/55/111&0.295 &0.277 \\
		    \cdashlinelr{1-7}
      2 & \multirow{2}{*}{Concat} &Raw  & 0.290 &28/21/124& 0.456 & 0.487  \\
		3&&+POS & 0.355 &52/21/100 & 0.507 & 0.517 \\
          \cdashlinelr{1-7}
		4&\multirow{2}{*}{HQE}&Raw & 0.345 &47/21/105 & 0.501 & 0.501 \\
		5&&+POS  & 0.353 &53/21/99 & 0.518 & 0.534\\
		    \cdashlinelr{1-7}
		6&\multirow{3}{*}{NTR}  & LSTM (+Atten.)  & 0.293  &26/42/105& 0.411 & 0.410 \\
         7&&QuReTeC & 0.367 &51/36/86 & 0.522 & 0.536 \\
		8&&T5 & \textbf{0.380}$^{1-6}$ &31/97/45 & \textbf{0.566}$^{1-7}$ & \textbf{0.593}$^{1-7}$ \\
	\bottomrule
	\end{tabular}
\end{table}

Table~\ref{tb:reranking} compares the results of passage re-ranking based on
different query reformulation methods.
Compared to the full ranking results, all query reformulation methods show improvements.
The re-ranking effectiveness of NTR (T5) approaches the manual queries, with 0.380 MAP and 0.566 NDCG@3.
Second to NTR (T5), NTR (QuReTec) yields 0.367 MAP and 0.522 NDCG@3, slightly higher than scores for HQE (+POS) and Concat (+POS). 
The re-ranking results indicate that among all CQR methods, NTR (T5) generates the best queries for the BERT re-ranker, perhaps because it generates queries that more resemble natural language queries, and are thus well-suited to the BERT re-ranker, which was trained on natural language queries.
This explains why NTR (QuReTec) shows slightly better R@1000 and MAP using BM25 retrieval, but NTR (T5) yields significantly better early precision in full ranking. 
Also, note that although HQE (+POS) outperforms Concat (+POS) in early-precision metrics (i.e., NDCG@3, NDCG@1), they show comparable MAP in this condition, which suggests that HQE outperforms Concat in full ranking (see Table~\ref{tb:full_ranking}) mainly due to the gains from first-stage BM25 retrieval.

\subsection{Fusion Analysis of Query Variations}
\label{subsec:fusion}

To better understand how query variations improve the retrieval effectiveness of conversational search, we conduct a thorough examination of query fusion in our multi-stage pipeline.
Specifically, we investigate two fusion methods:

\begin{enumerate}

\item \textbf{Late fusion}: As shown in Figure~\ref{fig:configuration}(a), we directly conduct reciprocal rank fusion~\cite{rrf} on the full ranking results from the two CQR methods.

\item \textbf{Early fusion}: Since late fusion in practice involves multiple passes of BERT re-ranking, which dominate system overhead (see Table~\ref{tb:overhead}), we propose an alternative:\ our previous experiments on re-ranking show that better first-stage retrieval improves overall full ranking results, and thus we propose fusing the first-stage ranking results from the two CQR methods and re-ranking the fused results using one of the CQR methods. This pipeline is shown in Figure~\ref{fig:configuration}(b).
\end{enumerate}

\begin{figure}[t!]
\centering
\begin{subfigure}[t]{.49\columnwidth}
    \includegraphics[width=\columnwidth]{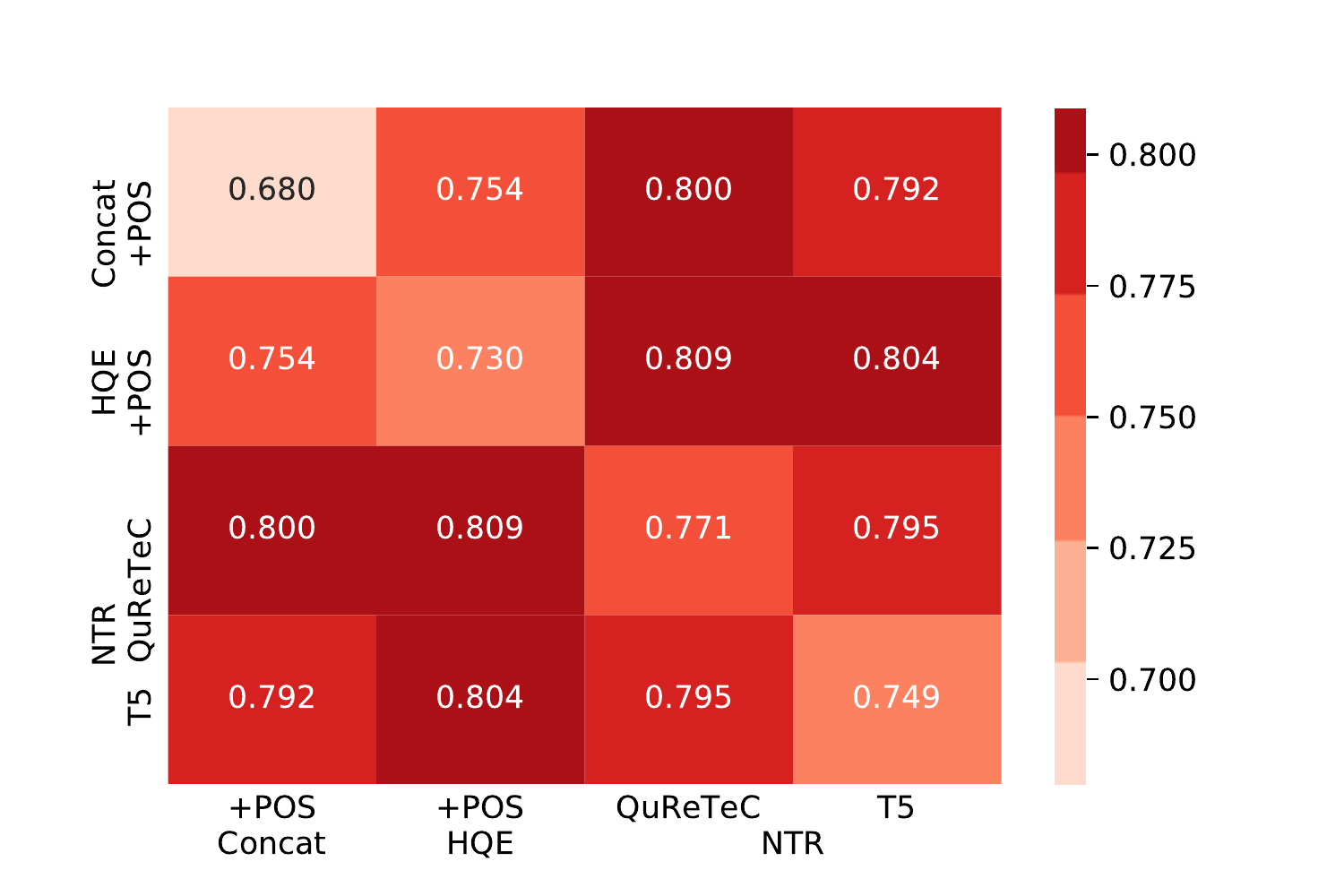}
    \caption{R@1000}
\end{subfigure}
\begin{subfigure}[t]{.49\columnwidth}
    \includegraphics[width=\columnwidth]{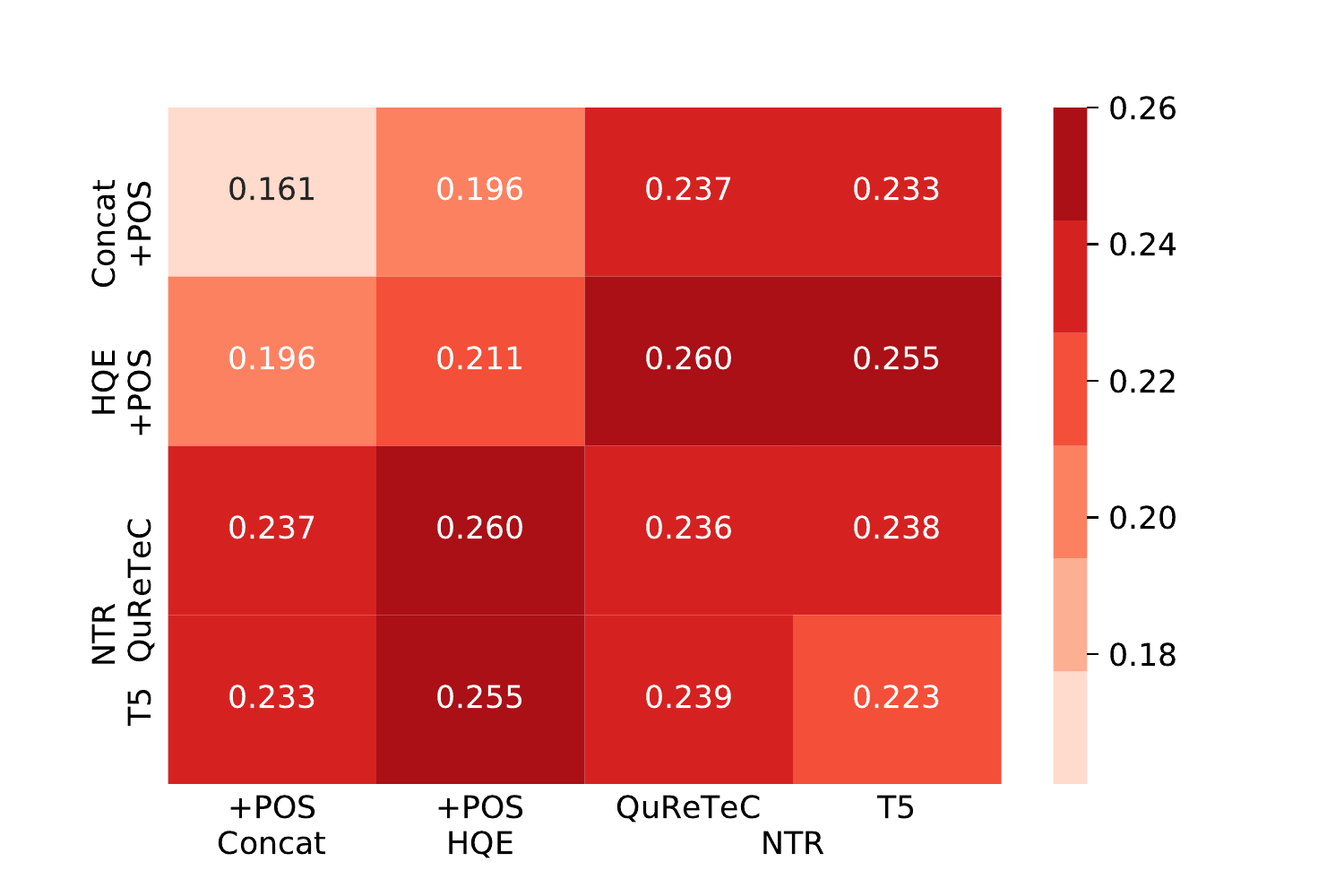}
    \caption{MAP}
\end{subfigure}
\caption{Grid search for first-stage (BM25) retrieval fusion.}
\label{fig:fusion}
\end{figure}

\noindent We first investigate whether first-stage retrieval can be improved by fusion.
Figure~\ref{fig:fusion} visualizes the outcomes of reciprocal rank fusion between selected pairs of CQR approaches at the BM25 retrieval stage. 
The best fusion results come from the combination of HQE (+POS) and NTR (QuReTeC or T5) in terms of R@1000 and MAP.
This is likely because, as observed previously, NTR and HQE improve conversational search in markedly different ways; fusing complementary results is often effective because it leverages different ranking features.
In contrast, fusion between QuReTeC and T5 leads to sub-optimal effectiveness.

\begin{table}[tb!]
	\caption{Fusion comparisons with HQE (+POS). Superscripts denote significant improvements ($p$-value $<0.05$) compared to the paired conditions.}
	\label{tb:fusion_full_rank}
	\centering
    \small
    \resizebox{\textwidth}{!}{
    \begin{tabular}{llllllllll}
    \toprule
       & & & \multicolumn{4}{c}{BM25}& \multicolumn{3}{c}{Full ranking (BM25+BERT re-ranking)}\\
        \cmidrule(lr){4-7} \cmidrule(lr){8-10}
	    & &Fusion method & R@1000 & MAP & NDCG@3 & NDCG@1 & MAP & NDCG@3 & NDCG@1 \\
	\midrule
      1&\multirow{2}{*}{NTR (QuReTec)} & Late & - &- & - & - & 0.367 & 0.535 & 0.540  \\
        2&& Early & \textbf{0.809} & \textbf{0.260} & 0.313 & 0.318 & 0.361 & 0.511 & 0.528  \\
       \cdashlinelr{1-10}
      3&\multirow{2}{*}{ NTR (T5)}  & Late & - & - &- & - & 0.371& 0.551 & 0.553\\
	  4& & Early & 0.804 & 0.255 & \textbf{0.325} & \textbf{0.337} & \textbf{0.375} &\textbf{0.565}$^{2}$ & \textbf{0.598}$^{1,2}$ \\
     
	\bottomrule
	\end{tabular}
	}
\end{table}

Since T5 and QuReTeC have comparable BM25 retrieval effectiveness when combined with HQE (+POS), we compare their fusion on full ranking.
Note that, in this experiment, we conduct early fusion using the queries generated by NTR for BERT re-ranking.
As shown in Table~\ref{tb:fusion_full_rank}, the two NTR methods can be further improved when fusing HQE (+POS); overall, NTR (T5) early fusion with HQE (+POS) yields the best effectiveness.
NTR (T5) early fusion with HQE (+POS) outperforms its NTR (QuReTec) counterpart, especially in terms of early precision (i.e., NDCG@3), despite having comparable effectiveness on fused BM25 retrieval. 
Note that whereas late fusion yields better full ranking accuracy for NTR (QuReTec), the reverse trend is seen with NTR (T5). 
This indicates that HQE (+POS) and NTR (QuReTec) yield BERT re-ranking results that can be further improved by fusion.
In contrast, NTR (T5) performs well at the BERT re-ranking stage, and late fusion provides no help, which is also consistent with our observations in Section~\ref{sec:reranking}. 
Thus, we advocate fusing the proposed CQR modules into state-of-the-art IR pipelines using early fusion.

\section{Query Variation Characteristics}\label{sec:discuss}

To further explore the distinct behaviors of HQE and NTR discovered in Section~\ref{sec:fullrank}, we present a study to analyze their differences from the following three perspectives:
\begin{enumerate}
    \item
    Query characteristics in terms of embeddings and textual content;
    \item
    Retrieval characteristics in terms of aggregation based on turn depth and sessions;
    \item
    Two case studies that illustrate the advantages and disadvantages of each model.
\end{enumerate}
Note that our analyses are based on the 20 sessions with relevance judgments in the TREC CAsT 2019 evaluation set.

\subsection{Intrinsic Characteristics: Query Representations}

An intuitive way to illustrate the characteristics of conversational queries is to visualize their embeddings using BERT.
This intuition, which comes from the MS MARCO conversational search task,\footnote{\url{https://github.com/microsoft/MSMARCO-Conversational-Search}}\saveFN\sft\ is based on the assumption that utterances in the same conversation session are similar in embedding space, as they are topic-oriented.

\begin{figure}[t!]
\centering
\begin{subfigure}[t]{.35\columnwidth}
    \includegraphics[width=\columnwidth]{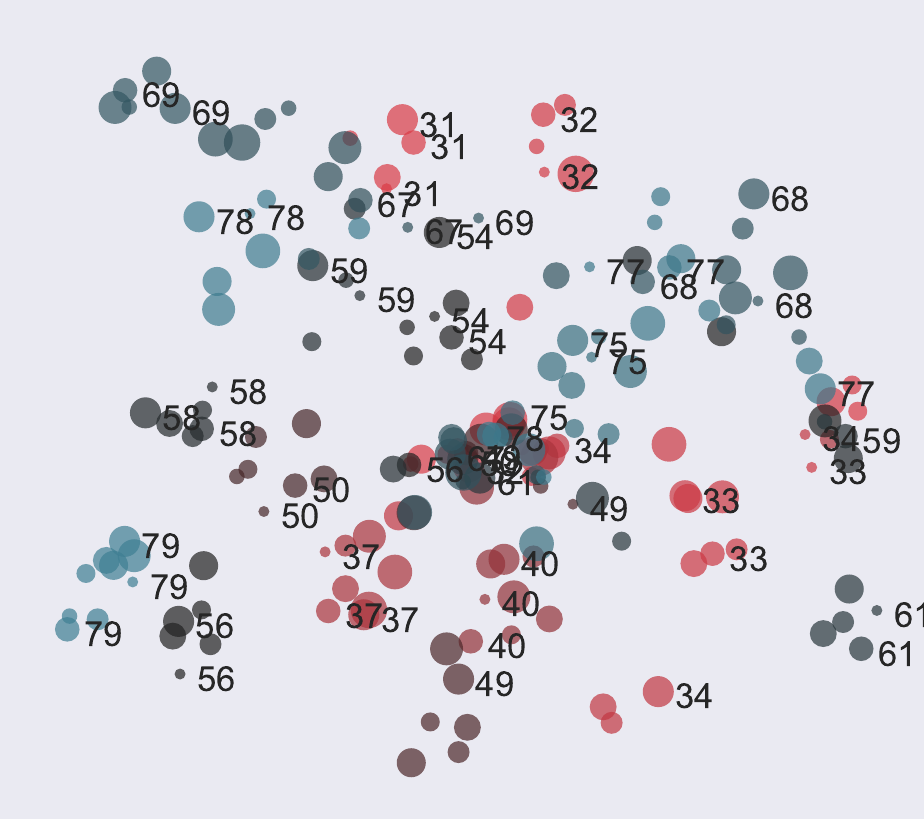}
    \caption{Raw query}
\end{subfigure}
\begin{subfigure}[t]{.35\columnwidth}
    \includegraphics[width=\columnwidth]{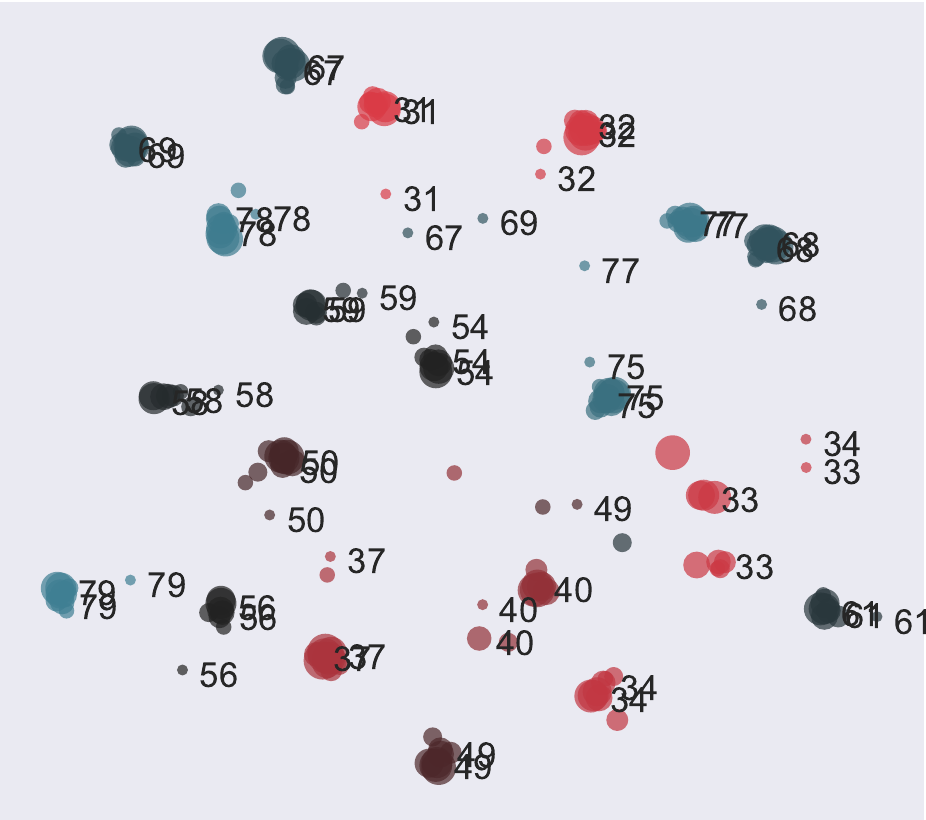}
    \caption{HQE (+POS)}
\end{subfigure}
\begin{subfigure}[t]{.35\columnwidth}
    \includegraphics[width=\columnwidth]{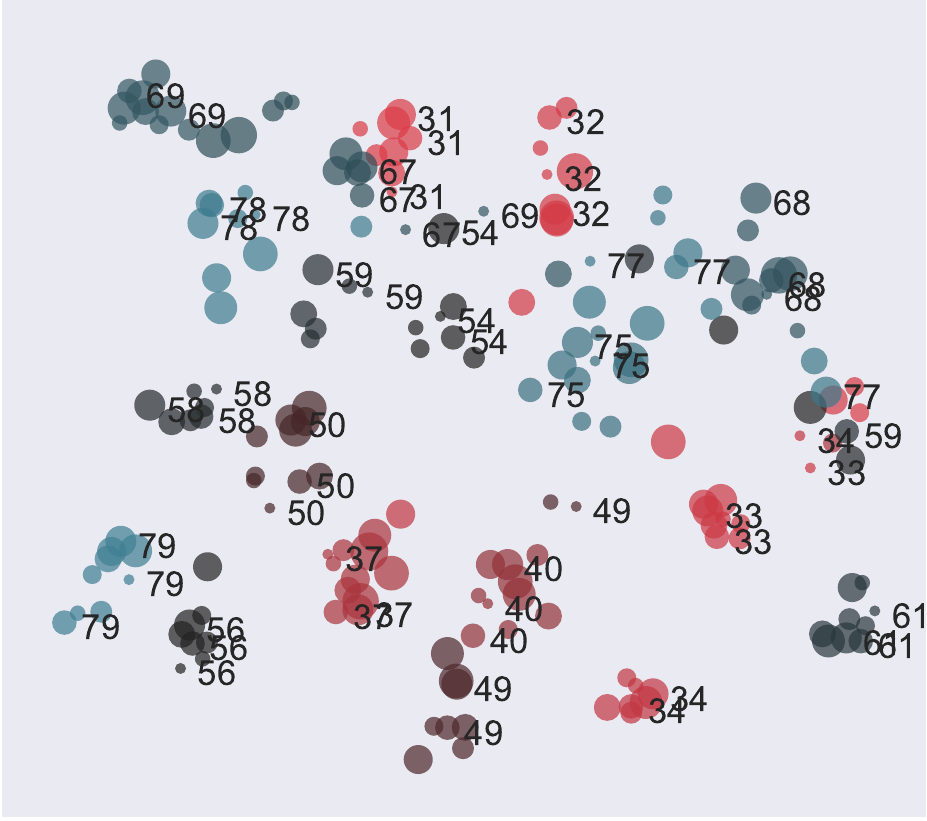}
    \caption{NTR (T5)}
\end{subfigure}
\begin{subfigure}[t]{.35\columnwidth}
    \includegraphics[width=\columnwidth]{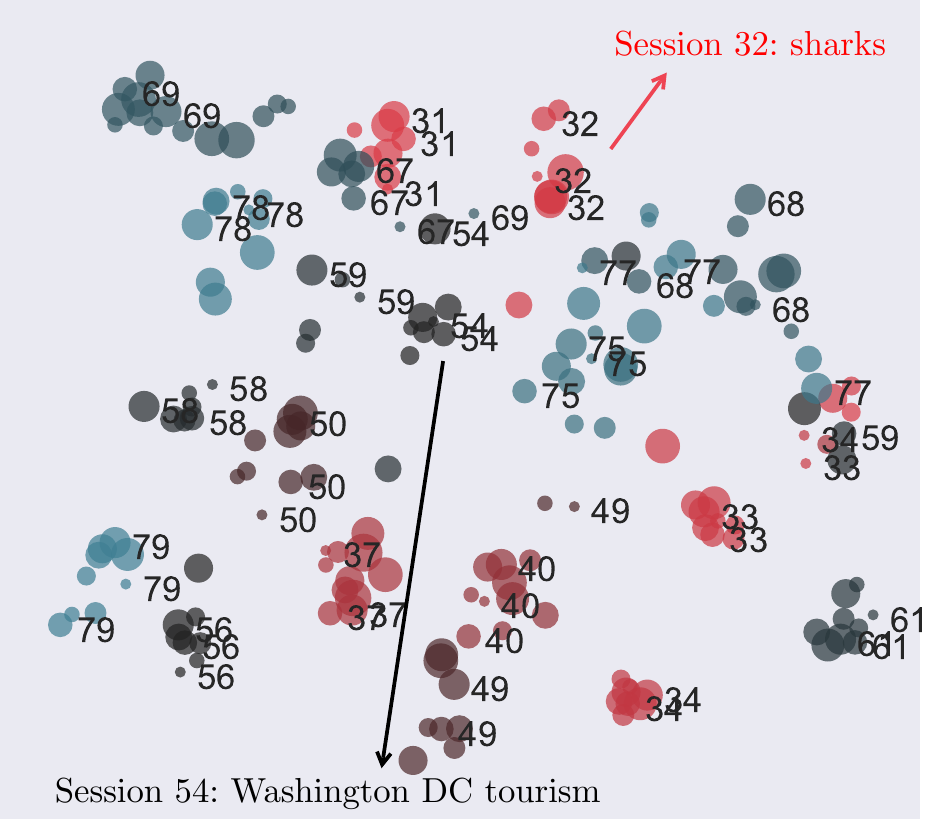}
    \caption{Manual}
\end{subfigure}
\caption{$t$-SNE plots of 20 sessions in the TREC CAsT 2019 evaluation set (colors for session IDs; dot sizes for turn depth), where the query embeddings are formed by average pooling over the second-to-last hidden layer of all tokens from BERT. Note that we apply $t$-SNE on the embeddings from the four types of queries together to ensure that they are in the same representation space.}
\label{fig:tsne_all_session}
\end{figure}

We leverage BERT~\cite{BERT} to project the queries---raw queries, HQE (+POS), NTR (T5), and manual queries---into embedding space and apply 2-dimensional $t$-SNE~\cite{tsne} on them together to ensure that they are in the same representation space.\footnote{Note that we follow the setup for building artificial conversational sessions from Bing search queries, see footnote~\useFN\sft\ for details.}
Panels (a)--(d) in Figure~\ref{fig:tsne_all_session} visualize their respective $t$-SNE embeddings, where color represents different session identifiers (IDs) and the embedding size reflects turn depth.

From Figure~\ref{fig:tsne_all_session}, we note the following:
First, the raw queries in panel~(a) show vague boundaries between sessions, especially those in the central region.
This may be due to ambiguity from coreferences and omissions in conversational utterances, as it is difficult to differentiate them without context.
Second, manual and NTR (T5) queries in panels~(c) and (d) form more well-defined clusters between sessions than raw queries, suggesting that they are more topically coherent.
Furthermore, observe that NTR (T5) and manual queries have similar embedding distributions, implying that the two models yield similar queries, explaining the many ``Ties'' in Tables~\ref{tb:full_ranking} and~\ref{tb:reranking}.
Finally, HQE (+POS) queries in panel (b) form well-defined clusters:\ queries in the same session overlap heavily, demonstrating that reformulations by HQE (+POS) in the same sessions are similar.

To further examine the high similarities between queries formulated by NTR (T5) and the manual queries, we measure query similarities quantitatively in terms of textual content.
Specifically, we compare query texts from different CQR methods with BLEU~\cite{BLEU}.\footnote{Using \texttt{multi-bleu-detok.perl} from~\citet{BLEU, CANARD}.}
Here, we take the manual queries as the reference sentences and calculate BLEU scores for the other methods.
As shown in Table~\ref{tb:bleu}, NTR (T5) queries yield the highest scores, and HQE (+POS) queries have the lowest.
Note that scores of the raw queries fall between those of NTR (T5) and HQE (+POS).
These results not only validate the high query similarities between NTR (T5) and manual queries observed in Figure~\ref{fig:tsne_all_session}, but also confirm that queries generated by HQE (+POS) are markedly different from the other methods.

\begin{table}[t!]
	\caption{BLEU scores with manual queries as references.}
	\label{tb:bleu}
	\centering
	\small
    \begin{tabular}{cccc}
	\toprule
	         & Raw queries  & HQE (+POS)  & NTR (T5) \\
	\midrule
	    BLEU & 60.41 & 33.73 & \textbf{76.22} \\
	\bottomrule
	\end{tabular}
\end{table}

\begin{figure}[t]
\begin{subfigure}[t]{.4\columnwidth}
    \includegraphics[width=\columnwidth]{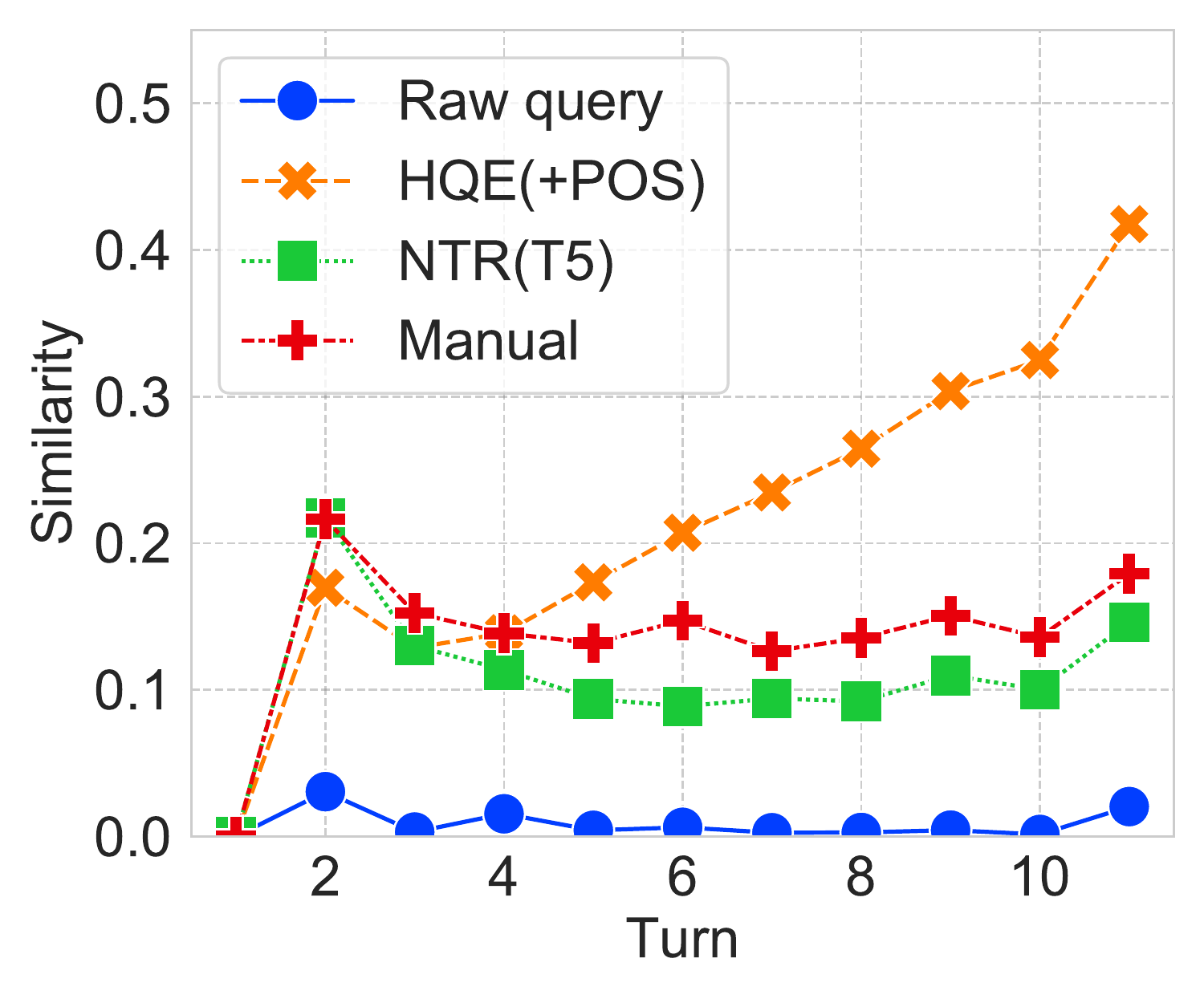}
    \caption{Turn similarity}
\end{subfigure}
\begin{subfigure}[t]{.4\columnwidth}
    \includegraphics[width=\columnwidth]{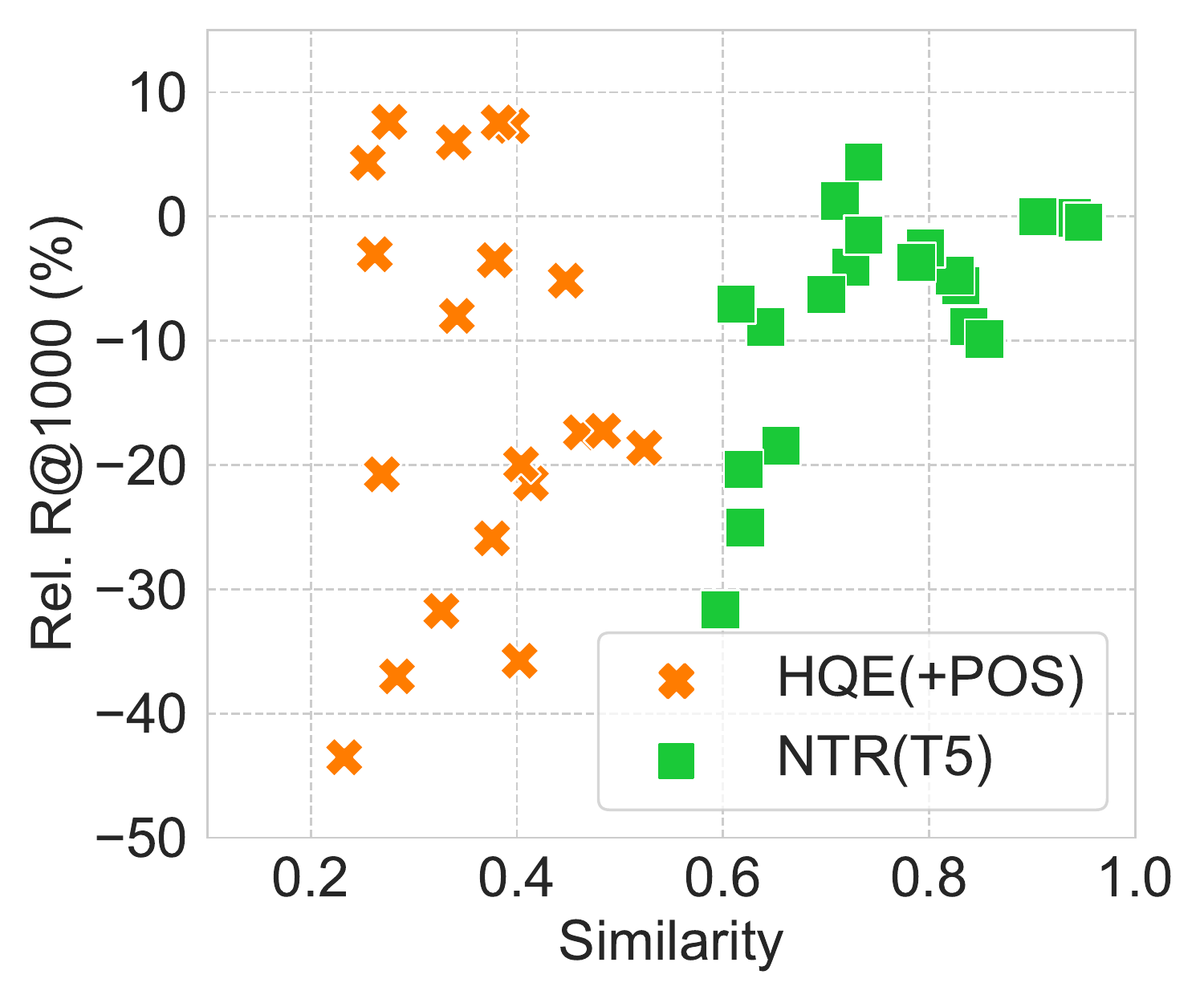}
    \caption{Session similarity}
\end{subfigure}
\caption{Analysis of results from first-stage retrieval. In panel (a), we compare the retrieved passages of each turn with results from the previous turn for different CQR methods using Jaccard similarity. In panel (b), we compare set similarities ($x$-axis) and BM25 retrieval effectiveness in terms of relative R@1000 ($y$-axis) of HQE (+POS) and NTR (T5) against manual queries.}
\label{fig:simialrity}
\end{figure}

\subsection{Extrinsic Characteristics: Retrieval Results}

Above, we clarify the distinct behaviors of the two CQR approaches from a query
perspective.
To further identify the reasons behind the ``Wins'' and ``Ties'' of HQE (+POS) and NTR (T5) versus manual queries in Table~\ref{tb:full_ranking}, we analyze the similarities of the retrieved sets from first-stage retrieval when different CQR methods are adopted.
In Figure~\ref{fig:simialrity}, the sets retrieved by BM25 are analyzed from the perspective of turn depth in panel~(a) and the perspective of sessions in panel~(b).
Specifically, we use Jaccard similarity $J(\cdot)$ to quantitatively analyze the retrieved sets.
In Figure~\ref{fig:simialrity}(a), the similarity for turn~$i$ is the averaged values of the Jaccard similarities between the~$(i-1)$-th and~$i$-th turns over all sessions. 
In Figure~\ref{fig:simialrity}(b), we take the retrieved sets from the manual queries as reference sets to calculate relative (rel.) R@1000 and $J(\cdot,\mathcal{P}_\text{Manual})$ of NTR (T5) and HQE (+POS).
Each pair of averaged metrics (rel.\ R@1000, $J(\cdot)$) over all turns in each session is plotted as a point in the figure.

We draw three conclusions from Figure~\ref{fig:simialrity}.
First, the ``Ties'' of manual queries and NTR (T5) in Table~\ref{tb:full_ranking} can be explained by the observations from panels~(a) and (b).
As shown in panel~(a), whereas the retrieved sets' similarities between NTR (T5) and manual queries remain around $0.15$ as the turns progress, the rel.\ R@1000 scores of NTR (T5) mostly cluster around $0$ on the $y$-axis in panel~(b).
Second, we conjecture that the ``Wins'' of HQE (+POS) in Table~\ref{tb:full_ranking} come from the upper-left cluster in Figure~\ref{fig:simialrity}(b);
this could be due to the disparate behaviors observed in Figure~\ref{fig:simialrity}(a): HQE (+POS) tends to retrieve similar sets as the turns progress.
Finally, Figure~\ref{fig:simialrity} illustrates not only a significant gap between HQE (+POS) and NTR (T5) in panel~(a) but also a clear boundary at around 0.55 on the $x$-axis in panel~(b).
These observations suggest that the success of the fusion approach (RRF) could be attributable to the different behaviors of these two methods, and we are exploiting complementary ranking signals.

\begin{table*}[t!]
\caption{Comparison of different queries for TREC CAsT 2019 session 32.}
\label{tb:case_study_session_32}
\centering
\small
\resizebox{\textwidth}{!}{
\begin{tabular}{p{0.04\textwidth}R{0.175\textwidth}R{0.18\textwidth}R{0.35\textwidth}R{0.175\textwidth}}
\toprule
Turn & Raw queries & Manual queries & HQE (+POS) & NTR (T5) \\
\midrule
\multirow{2}{*}{1--5} & \multicolumn{4}{c}{(\textit{We provide raw queries here as context}): (1) What are the different types of sharks?  (2) Are sharks endangered? If so, which species?} \\
  & \multicolumn{4}{c}{(3) Tell me more about tiger sharks. (4) What is the largest ever to have lived on Earth? (5) What's the biggest ever caught?} \\
\midrule
6 & What about for great whites? & What about for \textbf{great whites}? & \textbf{sharks} sharks \textbf{tiger} sharks largest Earth biggest  great whites What about for great whites? & What about for \textbf{great whites}? \\
\cdashline{1-5}
R@1000 & 0.177 & 0.177 & 0.824 & 0.177\\ 
\midrule
7 & Tell me about makos. & Tell me about \textbf{Mako sharks}. & sharks sharks tiger sharks largest Earth biggest makos  Tell me about makos. & Tell me about \textbf{makos}. \\
\cdashline{1-5}
R@1000 & 0.273  & 1.000 & 1.000 & 0.273\\ 
\midrule
8 & What are their adaptations? & What are \textbf{Mako shark adaptations}? & \textbf{sharks sharks tiger sharks} largest Earth biggest \textbf{makos adaptations}  What are their \textbf{adaptations}? & What are \textbf{makos adaptations}? \\
\cdashline{1-5}
R@1000 & 0.000 & 1.000 & 0.941  & 0.765\\
\bottomrule
\end{tabular}
}
\end{table*}

\begin{table*}[t!]
\caption{Comparison of different queries for TREC CAsT 2019 session 54.}
\label{tb:case_study_session_54}
\centering
\small
\resizebox{\textwidth}{!}{
\begin{tabular}{p{0.04\textwidth}R{0.175\textwidth}R{0.18\textwidth}R{0.35\textwidth}R{0.175\textwidth}}
\toprule
Turn & Raw queries & Manual queries & HQE (+POS) & NTR (T5) \\
\midrule
\multirow{2}{*}{1--4} & \multicolumn{4}{c}{(\textit{We provide raw queries here as context}): (1) What is worth seeing in Washington D.C.?  (2) Which Smithsonian museums are the most popular?} \\
  & \multicolumn{4}{c}{(3) Why is the National Air and Space Museum important? (4) Is the Spy Museum free?} \\
\midrule
5 & What is there to do in DC after the museums close? & What is there to do in Washington D.C. after the museums close? & worth Washington D.C. Smithsonian museums Space Museum Spy Museum DC museums  What is there to do in DC after the museums close? & What is there to do in DC after the \textbf{Smithsonian} museums close? \\
\cdashline{1-5}
R@1000 & 0.579 & 0.368 & 0.526 & 0.632\\ 
\midrule
6 & What is the best time to visit the reflecting pools? & What is the best time to visit the reflecting pools in Washington D.C.? & worth Washington D.C. \textbf{Smithsonian museums} \textbf{Space Museum} \textbf{Spy Museum} \textbf{DC museums} pools  What is the best time to visit the reflecting pools? & What is the best time to visit the reflecting pools of Washington D.C.? \\
\cdashline{1-5}
R@1000 & 0.250 & 1.000 & 0.000 & 1.000\\ 
\midrule
7 & Are there any famous foods? & Are there any famous foods in \textbf{Washington D.C.}? & worth Washington D.C. \textbf{Smithsonian museums} \textbf{Space Museum} \textbf{Spy Museum} \textbf{DC museums} pools famous foods  Are there any famous foods? & Are there any famous foods in \textbf{Washington D.C.}? \\
\cdashline{1-5}
R@1000 & 0.000  & 0.500 & 0.000 & 0.500\\
\bottomrule
\end{tabular}
}
\end{table*}

\subsection{Case Studies}

Tables~\ref{tb:case_study_session_32} and~\ref{tb:case_study_session_54} present examples from sessions~32 and 54 that showcase the advantages and disadvantages of HQE (+POS) and NTR (T5).
The row under each turn's queries shows the BM25 retrieval effectiveness (R@1000) of the four reformulation methods:\ raw queries, manual queries, HQE (+POS), and NTR (T5).\footnote{Due to space limitations, we only provide raw queries from earlier turns as context, for which HQE (+POS) and NTR (T5) have similar effectiveness.} 

Table~\ref{tb:case_study_session_32} compares the reformulated queries about sharks and shows that the queries reformulated by NTR (T5) lose the context word \textit{shark} after turn~6. Furthermore, from turns~7 to $10$, NTR (T5) considers the context to be \textit{makos} rather than \textit{makos shark};
hence, NTR (T5) is unlikely to retrieve passages with \textit{makos shark} compared to HQE (+POS) and manual queries. 
HQE (+POS) performs better in terms of R@1000 and ``Wins'', mainly due to the concatenation of the topic keyword \textit{shark}. 
In particular, in turn~6, HQE (+POS) greatly outperforms NTR (T5) and manual queries, the main reason being that the words \textit{great white} from NTR (T5) and manual queries guide the BM25 model to retrieve documents with both \textit{great} and \textit{white} but not relevant to \textit{shark}.
This example also demonstrates that manually-rewritten queries are not always the best.

However, HQE (+POS) can sometimes be too aggressive in injecting context into utterances.
As shown in Table~\ref{tb:case_study_session_54}, HQE (+POS) puts too much emphasis
on \textit{museum} when the subtopic changes to \textit{reflecting
pool} in turn~6 and \textit{food (D.C. half smoke)} in turn~7.
In contrast, NTR (T5) mimics humans in inserting adequate context in
utterances.
For instance, NTR (T5) puts \textit{Washington D.C.} in turn~7 as context for the BM25 model to understand the raw utterance.
Moreover, take turn~5 as an example:\ NTR (T5) sometimes fills in missing context omitted by the human annotators (in this case, adding the word \textit{Smithsonian}); for this reason, NTR (T5) outperform manual queries in a few cases.

\section{Component Evaluations}

Thus far, we have empirically analyzed our proposed multi-stage ConvPR pipeline with different CQR methods.
To better understand the advantages and disadvantages of HQE and NTR, we further conduct sensitivity analyses, ablation studies, and system overhead measurements.

\subsection{Sensitivity Analyses}\label{sec:sens}
We conduct a sensitivity analysis of HQE (+POS) and NTR (T5) on the TREC CAsT 2019 training set, with hyperparameters tuned based on BM25 retrieval effectiveness in terms of R@1000 and MAP. 

\begin{figure}[t!]
\centering
\begin{subfigure}[t]{.45\columnwidth}
    \includegraphics[width=\columnwidth]{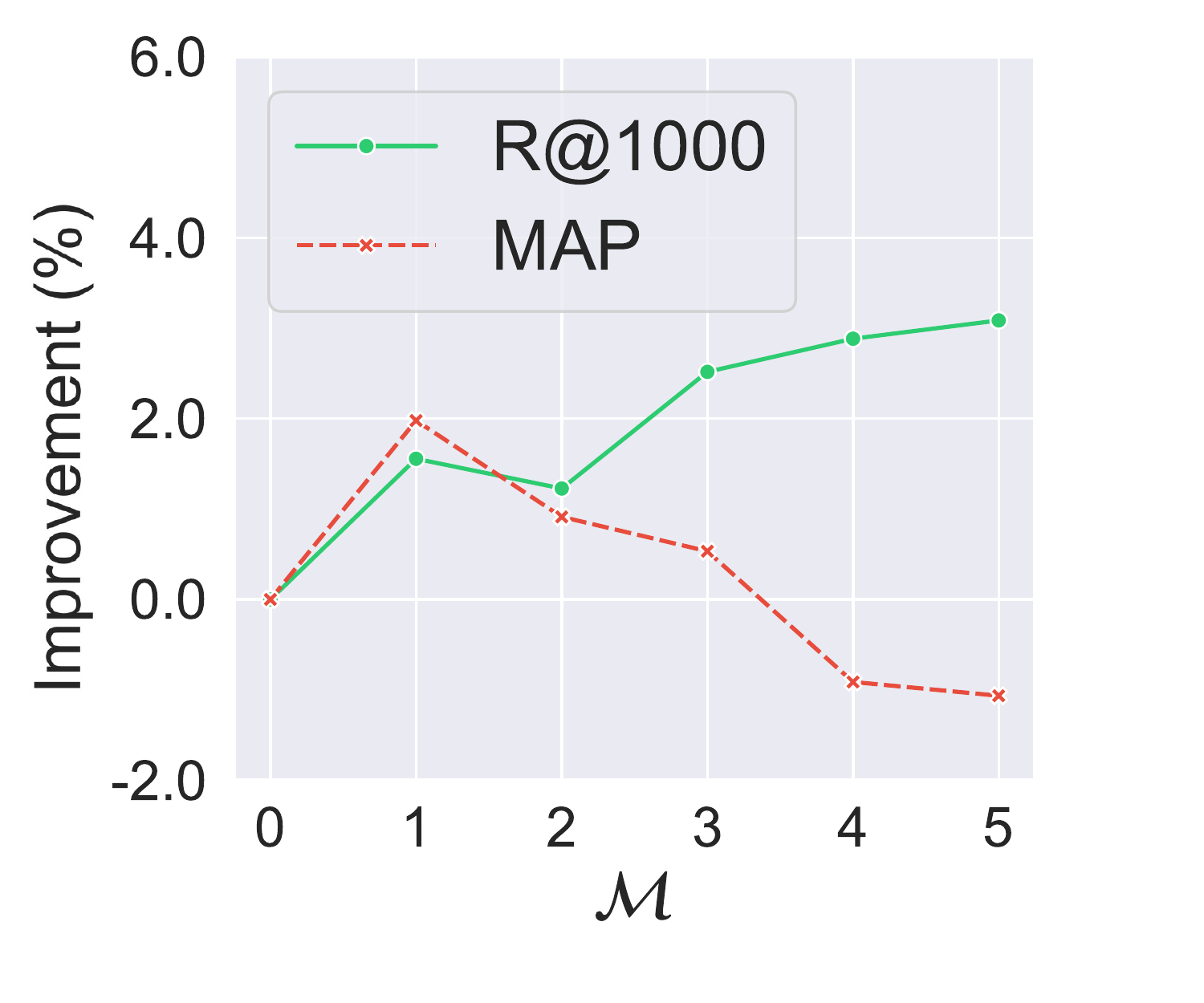}
    \caption{HQE (+POS)}
\end{subfigure}
\begin{subfigure}[t]{.45\columnwidth}
    \includegraphics[width=\columnwidth]{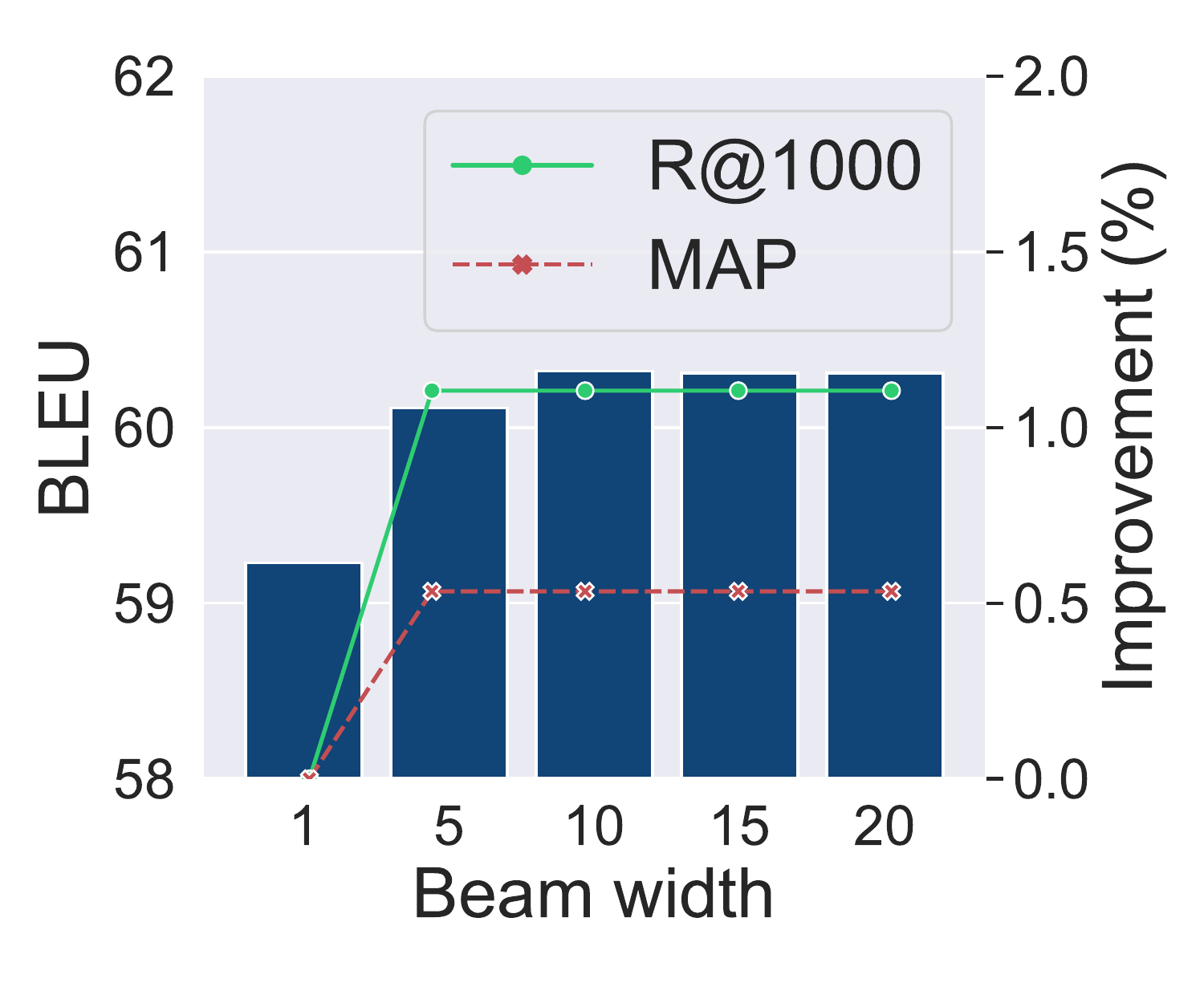}
    \caption{NTR (T5)}
\end{subfigure}
\caption{Sensitivity analyses. Panel (a) shows BM25 retrieval effectiveness sensitivity from adding subtopic keywords from the previous $\mathcal{M}$ turns for HQE (+POS). Panel (b) shows the sensitivity of beam width in beam search decoding in terms of BLEU scores (bars) and BM25 retrieval effectiveness (lines) for NTR (T5).}
\label{fig:sensitivity}
\end{figure}

\smallskip \noindent
\textbf{HQE (+POS).}
Figure~\ref{fig:sensitivity}(a) shows grid search results on R@1000 and MAP. 
Specifically, we tune $\mathcal{R}_{\text{topic}}$, $\mathcal{R}_{\text{sub}}$,
$\eta$, and $\mathcal{M}$ for optimal R@1000 and MAP separately.
By fixing $(\eta, \mathcal{R}_{\text{topic}}, \mathcal{R}_{\text{sub}})$ at the
best R@1000, $(10, 4.5, 3.5)$, and at the best MAP, $(12, 4.0, 3.0)$,
Figure~\ref{fig:sensitivity}(a) shows grid search results in terms of
various values of $\mathcal{M}$.

We first note from Figure~\ref{fig:sensitivity}(a) that both R@1000
and MAP improve when $\mathcal{M}>0$, indicating that adding subtopic keywords
from the previous $\mathcal{M}$ turns is effective for query expansion.
In addition, R@1000 and MAP yield different trends in the grid search, with the
best $\mathcal{M}=5$ and $\mathcal{M}=1$ for R@1000 and MAP, respectively, suggesting that the optimal query for BM25 search is different for each metric.
Thus, in the previous experiments, we generated HQE (and Concat) queries using
the hyperparameters with the best R@1000 for BM25 first-stage retrieval and the best MAP for BERT re-ranking.\footnote{For Concat, the best $\mathcal{M}$ is $9$. Due to the high computational costs of tuning hyperparameters for BERT re-ranking, we directly use the best value for BM25 search.}

\smallskip \noindent
\textbf{NTR (T5).}\label{sec:bleu_sens}
We also analyze the sensitivity of beam width~$w$ in beam search decoding for NTR (T5) in Figure~\ref{fig:sensitivity}(b), where bars denote the
BLEU scores (left $y$-axis)---used for evaluating machine-translated texts~\cite{BLEU} and also as a quality indicator in~\citet{CANARD}---and lines denote the improvements in IR metrics compared to beam width $w = 1$ (right $y$-axis).
Note that $w$ stands for the number of partial sequences with the highest probabilities the model keeps in order to find a single sequence with a limited-width breadth-first search during the decoding process.
To determine the optimal hyperparameters for CAsT query inference, we consider the development (dev) set in CANARD and the training set of CAsT to choose $w$ in the range of $\{1, 5, 10, 15, 20\}$.
Specifically, Figure~\ref{fig:sensitivity}(b) illustrates the BLEU score versus beam width on the CANARD dev set and R@1000 and MAP versus beam width on the CAsT training set.
Observe that $w = 10$, the best beam width, yields the highest BLEU
(60.32) on the CANARD dev set.\footnote{T5 achieves a better BLEU score than the 51.37 of LSTM (+Atten.) on the dev set of CANARD~\cite{CANARD}.}
As for the best beam width for retrieval effectiveness, $w = 5$ achieves R@1000 (+1.1 points) and MAP (+0.5 points) compared to $w = 1$ in the training set of CAsT.
To maintain query reformulation quality without hurting IR effectiveness, we choose $w = 10$ in all of our experiments.

\subsection{Ablation Studies}\label{sec:ablation}

\noindent
\textbf{HQE (+POS).}
To evaluate the effectiveness of each module in HQE (+POS), we turn off each module separately and test the reformulated queries at the BM25 retrieval stage, as shown in Table~\ref{tb:hqe_ablation}.
We first separately remove topic and subtopic keywords (conditions~2 and 3) from the HQE (+POS) queries.
Compared to the original HQE (+POS) queries (condition~1), removing topic keywords significantly degrades effectiveness in terms of all metrics.
On the other hand, only R@1000 suffers when removing subtopic keywords. 
In addition, we remove the QPP module in condition~4, under which, in contrast to condition~3, all subtopic keywords are used for query expansion (even if the query is not ambiguous).
The slight drop in R@1000 and MAP indicates that without QPP measurements, subtopic keywords result in the addition of noisy terms to the queries.

Furthermore, we are able to investigate the effects of HQE term weights, as our design assigns greater term weights to topic keywords compared to subtopic keywords.
To accomplish this, we remove the effects of the weights by assigning an equal weight to all the terms in each rewritten query.
The results shown in condition~5 indicate that our proposed HQE term weights have a positive impact on ranking effectiveness.

\begin{table}[t!]
	\caption{HQE (+POS) ablation study on BM25 retrieval.}
	\label{tb:hqe_ablation}
	\centering
	\small
    \begin{tabular}{ccccccccc}
	\toprule
	 Condition&  Topic & Subtopic & QPP & Term weight  & R@1000 & MAP  & NDCG@3 & NDCG@1   \\
	\midrule
    1&\checkmark  &\checkmark &\checkmark&\checkmark & 0.730  &0.211 &0.259 &0.261 \\
    2& &\checkmark&\checkmark &\checkmark &0.550  &0.144 &0.187 &0.200 \\
	3& \checkmark& &\checkmark&\checkmark & 0.714&0.207& 0.259 & 0.260 \\
     4&\checkmark &\checkmark& &\checkmark & 0.728 &0.207 & 0.264 & 0.263  \\
     5&\checkmark &\checkmark&\checkmark & & 0.714 &0.170 & 0.234 & 0.242 \\
     
	\bottomrule
	\end{tabular}
\end{table}

\smallskip \noindent
\textbf{NTR (T5).}
Since the NTR (T5) method transfers query rewriting knowledge from the CANARD dataset, we wish to examine differences between in-domain and out-of-domain training data when comparing NTR (T5) to HQE (+POS).
Here, we use two kinds of human annotations as CQR supervision signals:\ (1)~annotated queries in the TREC CAsT 2019 evaluation set, and (2)~the CANARD training set as discussed in Section~\ref{sec:canard}.
Specifically, when training the CQR model with the CAsT annotations, we use only those queries in the 30 out of 50 sessions without relevance labels (see Section~\ref{sec:cast}).
Furthermore, we also ablate the pretrained weights to determine whether the CQR model can be trained from scratch.
Observe from conditions~1 and 2 of Table~\ref{tb:t5_ablation}:\ although NTR (T5) indeed benefits from the out-of-domain CANARD data (+0.014 in R@1000 compared to condition~2), it still demonstrates competitive retrieval effectiveness when using only the limited number of instances from the CAsT dataset:\ same R@1000 but higher scores in ranking metrics MAP/NDCG@3/NDCG@1 0.221/0.277/0.278 compared to condition~1 in Table~\ref{tb:hqe_ablation}, MAP/NDCG@3/NDCG@1 0.211/0.259/0.261.
These results show that, surprisingly, only a modest amount of training data is sufficient to learn a good CQR model.

When comparing conditions~3 and 4, we observe that the CANARD-trained CQR model without pretrained weights is still better than the model that learns from only the CAsT dataset (+0.048 in R@1000).
From Table~\ref{tb:t5_ablation}, we conclude that for the CQR task, linguistic knowledge inherited from large-scale pretraining is necessary to achieve good retrieval effectiveness.

\begin{table}[t!]
	\caption{NTR (T5) ablation study on BM25 retrieval.}
	\label{tb:t5_ablation}
	\centering
    \small
    \begin{tabular}{cclcccc}
    \toprule
	    Condition & Pretraining & Fine-tuning & R@1000 & MAP & NDCG@3 & NDCG@1\\
	\midrule
        1 & \checkmark & CANARD & 0.744 & 0.223 & 0.295 & 0.293 \\
        2 & \checkmark & CAsT   & 0.730 & 0.221 & 0.277 & 0.278 \\
 	\cdashline{1-7}
 	    3 & & CANARD & 0.049 & 0.008 & 0.010 & 0.009 \\
        4 & & CAsT   & 0.001 & 0.000 & 0.000 & 0.000 \\
	\bottomrule
	\end{tabular}
\end{table}

\subsection{System Overhead Measurements}

In addition to retrieval effectiveness, we further discuss the efficiency of our multi-stage pipeline.
Table~\ref{tb:overhead} shows measurements of system overhead in our pipeline.
For NTR (T5), we measure latency on a Google Colab TPU with a fixed input sequence of length 512 and a batch size of 8.\footnote{As each TPU (TPUv2) has 8 cores, the minimum batch size should be 8. For a fair comparison, we calculate system overhead as total time $\times$ 8 / \# of queries.}
For HQE (+POS) and BM25 search, we measure the CPU overhead on a single thread. 
With a small beam width, $w=1$, the TPU inference latency of NTR (T5) is comparable to the non-neural HQE (+POS) model.
However, as the beam width of the decoding process increases, the inference time for NTR (T5) grows significantly.
Setting the beam width over 5 can slightly improve retrieval effectiveness in terms of R@1000 (+1\%) and MAP (+0.5\%), see Figure~\ref{fig:sensitivity}(b).
However, considering both effectiveness and efficiency, $w=1$ seems like a good setting for NTR (T5).
As for ranking itself, compared to BM25, the BERT re-ranker consumes considerable overhead using the TPU, which highlights the efficiency advantage of our proposed early fusion technique over late fusion.

\begin{table}[t!]
	\caption{System overhead. Time measured in seconds per query; $w$ represents the beam width for NTR (T5) beam search.}
	\label{tb:overhead}
	\centering
	\small
    \begin{tabular}{llcc}
	\toprule
	  \multicolumn{2}{l}{Module}  &TPU (sec/query)  & CPU (sec/query) \\
\midrule
   \multicolumn{2}{l}{HQE (+POS)}    & - & 0.045  \\
  \cdashlinelr{1-4}
  \multirow{3}{*}{NTR (T5)} & $w$=1 & 0.066 & -  \\
   &  $w$=5 & 0.613 & -  \\
   & $w$=10 & 1.464 & -  \\
\midrule
  \multicolumn{2}{l}{BM25 search} & - & 0.155  \\
  \midrule
  \multicolumn{2}{l}{BERT  re-ranking~\cite{doctttttquery}}  & 3.290 & -  \\
	\bottomrule
	\end{tabular}
\end{table}

\section{Future Work and Conclusions}

In this paper, we demonstrate that conversational query reformulation (CQR) combined with existing IR systems for standalone queries effectively addresses the challenging task of conversational passage retrieval (ConvPR). 
To tackle coreference and omission problems in conversational queries, we propose two CQR methods, Historical Query Expansion (HQE) and Neural Transfer Reformulation (NTR). 
Inspired by term importance estimation in IR, we propose HQE to expand a conversational query using important terms extracted from its conversational context. 
From the perspective of conversational query understanding, NTR, built on the pretrained sequence-to-sequence model T5~\cite{t5}, directly reformulates a conversational query into one that is natural for human
understanding.
Our empirical results on the CAsT benchmark dataset from TREC 2019 show that the two proposed CQR methods significantly improve output quality both for the BM25 retrieval and BERT re-ranking stages.
Furthermore, while NTR outperforms the best submission (a variant of HQE) in TREC CAsT 2019 by 18\% in terms of NDCG@3, the fusion of the two CQR methods yields even better results:\ a 23\% improvement in NDCG@3 compared to the best submission in TREC CAsT 2019. 

In addition to the promising results from CQR, our analyses also provide insights on the advantages and disadvantages of the two proposed approaches. 
Our case studies demonstrate HQE's advantage in expanding queries using topic-related words.
However, there is still a considerable effectiveness gap between HQE and manual queries, as HQE is built on a traditional bag-of-words model (BM25) that lacks understanding of conversational context.
On the other hand, the comparison with manual queries shows NTR's superior capability to mimic the way humans rewrite conversational queries.
Nevertheless, human-rewritten queries are not necessarily the gold standard for IR systems, as these queries sometimes omit important keywords.

Finally, the distinctive characteristics of our CQR methods contribute to the success of our final rank fusion module, which combines the ranked lists obtained from HQE and NTR, and further advances the state-of-the-art retrieval effectiveness.
More importantly, this finding paves the way for future work:\ directly combining HQE and NTR into a unified CQR module to generate better standalone queries for IR systems.
One possible way to accomplish this is to expand queries using dialogue context for NTR reformulated queries. 
The potential research questions include:\ (1) when to expand queries, and (2) how to weight query expansion terms.
In addition, despite the success of multi-stage retrieval systems, the inclusion of a CQR module increases the already-expensive computational costs, making it infeasible for some real-world applications.
Thus, another line of future work is to combine CQR and IR modules into an efficient unified model.
There remains much more work to be done in advancing the field towards intelligent agents capable of supporting conversational search.

\section{Acknowledgments}

This research was supported in part by the Canada First Research Excellence Fund, the Natural Sciences and Engineering Research Council (NSERC) of Canada, and the Ministry of Science and Technology in Taiwan under grant MOST 107-2218-E-002-061.
Additionally, we would like to thank Google for supporting this work by providing Google Cloud credits via the TensorFlow Research Cloud program.
Finally, we are grateful to the anonymous reviewers whose insightful comments from have helped to improve this work.

\bibliographystyle{ACM-Reference-Format}
\bibliography{main}

\end{document}